\documentclass[10pt,twocolumn,letterpaper]{article}

\usepackage{iccv}
\usepackage{times}
\usepackage{epsfig}
\usepackage{amsmath}
\usepackage{amssymb}
\usepackage{booktabs}
\usepackage{multirow}
\usepackage{graphicx}
\usepackage[title]{appendix}

\usepackage[breaklinks=true,bookmarks=false,colorlinks=True]{hyperref}

\iccvfinalcopy

\ificcvfinal\fi

\begin{document}

\title{Generative Adversarial Registration for Improved Conditional Deformable Templates}

\author{Neel Dey\\
New York University\\
{\tt\small neel.dey@nyu.edu}
\and
Mengwei Ren\\
New York University\\
{\tt\small mengwei.ren@nyu.edu}
\and
Adrian V. Dalca\\
MIT, MGH\\
{\tt\small adalca@mit.edu}
\and
Guido Gerig\\
New York University\\
{\tt\small gerig@nyu.edu}
}

\maketitle
\ificcvfinal\thispagestyle{empty}\fi

\begin{abstract}
   Deformable templates are essential to large-scale medical image registration, segmentation, and population analysis.
   Current conventional and deep network-based methods for template construction use only regularized registration objectives and often yield templates with blurry and/or anatomically implausible appearance, confounding downstream biomedical interpretation. We reformulate deformable registration and conditional template estimation as an adversarial game wherein we encourage realism in the \textit{moved} templates with a generative adversarial registration framework conditioned on flexible image covariates. 
   The resulting templates exhibit significant gain in specificity to attributes such as age and disease, better fit underlying group-wise spatiotemporal trends, and achieve improved sharpness and centrality. These improvements enable more accurate population modeling with diverse covariates for standardized downstream analyses and easier anatomical delineation for structures of interest.
\end{abstract}

\section{Introduction} \label{sec:intro}
Deformable image registration enables the quantification of geometric dissimilarity via the pairwise warping of a source image to a target. In the context of population studies, pairwise registration of a subject onto a \textit{deformable template} is a central step in standardized analyses, where an ideal template is an unbiased barycentric representation of the (sub-)population of interest~\cite{avants2004geodesic,avants2010optimal,joshi2004unbiased}. Templates play a key role in diverse large-scale biomedical imaging tasks such as alignment to a common coordinate system~\cite{fonov2011unbiased,talairach19883}, brain extraction~\cite{han2018brain,iglesias2015multi}, segmentation~\cite{cabezas2011review,iglesias2015multi}, and image and shape regression models~\cite{fishbaugh2013shaperegression,niethammer2011geodesic}, among others. 

While templates can be obtained from a reference database, they are preferably constructed for specific populations by optimizing for an image which minimizes the average deformation to each individual subject. As the template strongly affects subsequent morphometric analysis~\cite{senjem2005comparison,thompson2000mathematical}, template construction has received significant attention.
Further, as a single template cannot capture the wide structural variability within a population (via age and cohort, for example), we consider \textit{conditional} template estimation with continuous and/or categorical attributes. Conditional templates constructed on image sets with diverse covariates enable sub-population modeling accounting for information learned from the overall population and obviate the need for arbitrary thresholding of demographic information to perform independent analyses~\cite{cheng2020surface,dalca2019learning,yu2020mlmi}.

Implicit models for template estimation~\cite{allassonniere2007towards,avants2004geodesic,avants2010optimal,joshi2004unbiased,lorenzen2006media,wu2011sharpmean} alternate between registration of each scan to the current template estimate and updating the template based on averages of the warped subject scans. 
Due to the averaging of aligned image intensities, the resulting templates may blur significantly in regions with high-frequency deformations even alongside shape corrections~\cite{avants2010optimal}. Recently, \textit{explicit} template estimation via unsupervised deep networks was proposed in~\cite{dalca2019learning} where each stochastic update of a registration network yields a (potentially conditional) template without averaging aligned images and transformations.

However, both implicit and explicit models typically only minimize an image dissimilarity term between the moved template and fixed image (and/or vice-versa) subject to application-specific regularization ensuring a diffeomorphic (smooth, differentiable, and invertible) transformation. As inter-brain variability includes complex topological changes not captured by purely diffeomorphic models, estimated templates are often unrealistic and do not resemble the data that they represent.
Sub-optimal appearance impacts downstream applications due to ambiguous and/or implausible anatomical boundaries. For example, in order to register one or more expert-annotated templates to target images for atlas-based segmentation~\cite{iglesias2015multi,lee2019few,makropoulos2014automatic}, the template(s) must have clearly distinguishable anatomical boundaries to enable expert delineation. Unfortunately, structural anatomical boundaries are often obfuscated by current template estimation approaches.

We present a learning framework to estimate sharp (optionally conditional) templates with realistic anatomy via generative adversarial learning. Our core insight is that in addition to possessing high registration accuracy, the \textit{distribution} of moved template images should be indistinguishable from the real image distribution.
We develop a generator comprising of template generation and registration sub-networks and a discriminator which assesses the realism and condition-specificity of the synthesized and warped templates.
As adversarial objectives encourage high-frequency detail, the templates gain naturalistic boundaries without the need for ad hoc post-processing. To develop stable and accurate 3D GANs for large medical volumes with highly limited sample and batch sizes, we develop extensive optimization and architectural schemes, augmentation strategies, and conditioning mechanisms.

Our contributions include: (1) a generative adversarial approach to deformable template generation and registration which for the first time uses a realism-based registration regularizer;
(2) construction of conditional templates across diverse challenging datasets including neuroimages of pre-term and term-born newborns, adults with and without Huntington's disease, and real-world face images; (3) improvements on current template construction methodologies in terms of centrality and interpretability alongside significantly increased condition-specificity.
Code is available at \url{https://github.com/neel-dey/Atlas-GAN}.

\section{Related work} \label{sec:rel_work}

\textbf{Generative adversarial networks}~\cite{goodfellow2014generative}  have lead to remarkable progress in high-fidelity image generation~\cite{brock2018large,karras2019style,karras2020analyzing,Park_2019_CVPR}. Consequently, GANs for image translation~\cite{huang2018multimodal,isola2017image,Park_2019_CVPR,wang2018pix2pixHD,zhu2017unpaired} and inverse problems~\cite{daras2021intermediate,ledig2017cvpr,seitzer2018miccai} have shown that in addition to reconstruction objectives, adversarial regularizers dramatically increase the visual fidelity of the reconstructions by compensating for high-frequency details typically lost by using reconstruction objectives alone. We apply analogous reasoning in our use of conditional adversarial regularization of registration objectives. For conditional generator networks, modulating every feature map with learned conditional scales and shifts has lead to significantly improved image synthesis~\cite{brock2018large,park2019semantic,ren2021harmonization} over methods where the attribute vector is concatenated to the input.

\textbf{Deformable image registration} is the spatial deformation of a source image to a target. Optical flow registration commonly deployed in computer vision~\cite{brox2011pami,papenberg2006ijcv} admits deformations which may create anatomically implausible transformations when applied to biomedical images. Instead, a series of more suitable registration algorithms have been developed~\cite{christensen1997tmi,dinggang2002hammer,johnson2001ipmi,rueckert1999nonrigid,thirion1998image}, further leading to several topology-preserving diffeomorphic extensions~\cite{avants2008symmetric,beg2005computing,cao2005large,tustison2012ffd,VERCAUTEREN2009S61,zhang2019fast,shen2019region}. More recently, deep networks trained under either supervised~\cite{cao2017deformable,sokooti2017nonrigid,yang2017quicksilver} or unsupervised~\cite{balakrishnan2019vxm,dalca2019diffeo,de2017end,krebs2019learning,li2017nonrigid,niethammer2019metric} registration objectives have emerged, simultaneously offering both greater modeling flexibility and accelerated inference performance.

\textbf{Generative adversarial registration} leveraging simulation has been used in works such as~\cite{hu2018adversarial} where large-scale finite element simulations of plausible deformations serve as the real domain for a GAN loss alongside supervised registration. Simulated pairs of aligned and mis-aligned image patches have also been used to adversarially optimize a registration network~\cite{fan2019adversarial,fu2020lungregnet}. Our approach is distinct in that we focus on templates and not just registration, we develop adversarial registration techniques accounting for covariates, we process complete 3D volumes and do not use simulation, focusing only on moved template realism. 

\textbf{Template estimation} enables standardized analyses of image sets by acting as barycentric representations of a given population. Unconditional template construction has a rich history in medical image analysis~\cite{allassonniere2007towards,avants2004geodesic,avants2010optimal,joshi2004unbiased,lorenzen2006media,wu2011sharpmean}. Due to the blurring induced by image and shape averaging of aligned images, popular registration frameworks perform template post-processing and sharpening~\cite{avants2010optimal} which may inadvertently create implausible structure~\cite{antsx} and may still fail to resolve structures in highly variable populations. Further, given covariates of interest, ad hoc approaches may ignore shared information by dividing the dataset into sub-populations of interest and constructing templates for each independently. More principled approaches 
explicitly account for age and potentially other covariates by building spatiotemporal templates and have been extensively validated on pediatric~\cite{fishbaugh2019mfca,Gholipour2017,HABAS2010460,kim2020framework,schuh2018unbiased,SERAG20122255} and adult~\cite{Bne2020,davis2010population,huizinga2018neuro,schiratti2015learning} neuroimages. 

In this work, we extensively build upon VoxelMorph Templates~\cite{dalca2019learning} (referred to as \verb|VXM| in this work). Driven by a generative model, unconditional \verb|VXM| considers a grid of free parameters as a template, which is used together with a training image as input to a registration network~\cite{dalca2019diffeo}. The network estimates a diffeomorphic displacement field between each image and the template. Both the registration network and the template parameters are trained end-to-end under a regularized image matching cost. For conditional \verb|VXM|, a convolutional decoder upsamples an attribute vector to generate a conditional template, which is then similarly end-to-end processed by the registration network. Subsequent sections detail our methodologies and improvements.

\begin{figure*}[!t]
    \centering
    \includegraphics[width=\textwidth]{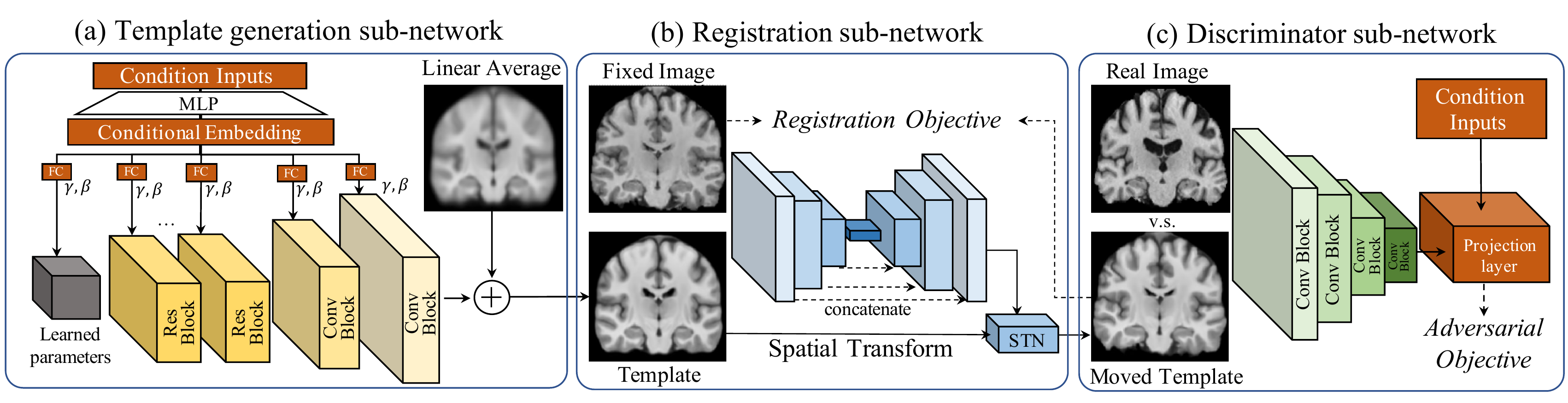}
    \caption{\textbf{Overview of the proposed template construction framework}. A template generation network (a) processes an array of learned parameters with a convolutional decoder whose feature-wise affine parameters are learned from input conditions to generate a conditional template. A registration network (b) warps the generated template to a randomly sampled fixed image. A discriminator (c) is trained to distinguish between moved synthesized templates and randomly-sampled images such that realism and condition-specificity is encouraged.}
    \label{fig:pipeline}
\end{figure*}

\section{Methodology}
Figure ~\ref{fig:pipeline} gives an overview of our approach. The generator network (a) \& (b) synthesizes a conditional template and deforms it to a fixed image to be assessed by a discriminator (c). The framework is trained
end-to-end under a regularized registration and adversarial cost to encourage both registration accuracy and template realism. 

\textbf{Template Generation Sub-network.} We develop an architecture whose backbone is agnostic to conditional or unconditional training. For unconditional training, we use a randomly-initialized parameter array (similar to~\cite{karras2019style,karras2020analyzing}) at half the spatial resolution of the desired template which is processed by 
convolutional decoder. The decoder output is added to the linear average of training images to generate the unconditional template, such that the network primarily learns to generate high-frequency detail. Checkerboard patterns generated by unconditional \verb|VXM| are ameliorated in this design by imposing spatial priors through convolutions. However, the central advantage of this backbone architecture is that it enables more parameter-efficient and powerful mechanisms for conditional training, as described below.

For conditional training, given condition vector $z$ and a feature map $h_{c}^{i}$ from the $i$th layer and $c$th channel, we feature-wise linearly modulate (FiLM~\cite{perez2018film}) all features $h_{c}^{i}$ in the backbone such that $FiLM(h_{c}^{i}) = \gamma_{c}^{i}(z) \ast h_{c}^{i} + \beta_{c}^{i}(z)$,
where $\gamma(z)$ and $\beta(z)$ are scale and shift parameters
learned from $z$. As shown in Figure~\ref{fig:pipeline}(a), we use a four-layer MLP to generate a shared conditional embedding from $z$ which is then linearly projected with weight decay individually to every layer in the template network to generate feature-wise transformations. 
The primary benefit of this design is that with conditioning at every layer (as opposed to conditional \verb|VXM| where the only source of conditioning is at its input), the template network has a higher capacity to fit datasets with high variability and synthesize more appropriate templates. A secondary benefit built upon the original \verb|VXM| architecture is parameter-efficiency. The original \verb|VXM| design uses a projection from $z$ to a high-dimensional vector at its input. In its neuroimaging experiments, $z \in \mathbb{R}^3$ (i.e., 3 attributes) is projected to a $\mathbb{R}^{\sim7M}$ vector 
using a weight matrix with $\sim 21M$ parameters. As the number of conditions increase (with one-hot encoded categorical attributes, e.g.), the rapidly increasing number of parameters in this weight matrix makes learning intractable. Conversely, our framework is relatively insensitive to the dimensionality of the condition vector $z$, which is processed by a shallow MLP (with 64 units) to generate channel-wise scalars.

\begin{figure*}[!ht]
    \centering
    \includegraphics[width=\textwidth]{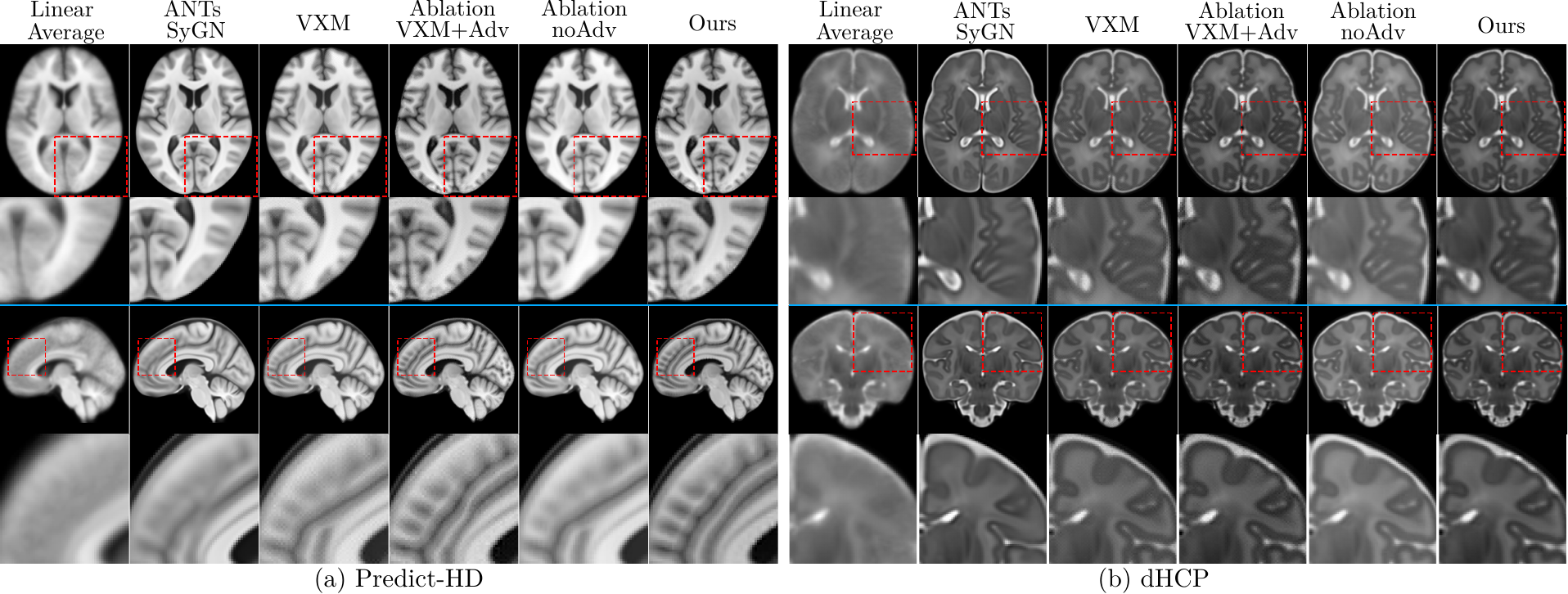}
    \caption{\textbf{Unconditional templates} learned from (a) Predict-HD and (b) dHCP. Our synthesized templates yield more neuroanatomically-representative structure (e.g., improved cortical folding for Predict-HD) and appearance (e.g., darker cortical grey matter for dHCP).
    }
    \label{fig:uncond-atlas}
\end{figure*}

\textbf{Registration Sub-network.} 
We use an established U-Net registration architecture~\cite{dalca2019diffeo}, which takes fixed and template images and outputs a time stationary velocity field (SVF) $v$~\cite{arsigny2006logeuc,ASHBURNER200795,modat2012pSVF}. When the SVF is integrated over time $t \in \left[ 0, 1 \right]$, it yields a diffeomorphic displacement field $\varphi_v^{(t)}$ such that $\frac{\partial \varphi_v^{(1)}}{\partial t} = v(\varphi_v^{(t)})$, where $\varphi_v^{(0)}$ and $\varphi_v^{(1)}$ represent the identity and final displacement fields, respectively. 
We then use $\varphi_v^{(1)}$ with a spatial transformer~\cite{jaderberg2015spatial} to deform the template to the fixed image space.

\textbf{Discriminator Sub-network.}
We use a five-layer fully convolutional discriminator network (PatchGAN~\cite{isola2017image,zhu2017unpaired}) to assess realism on local patches of input images. For example, given an input neuroimage volume of $160 \times 192 \times 160$, the discriminator has a receptive field of $63 \times 63 \times 63$. We find that discriminator regularization is essential for stable and artefact-free training, as outlined further below. 

For conditional templates, the discriminator is trained to distinguish between real and synthesized images given their categorical and/or continuous covariates. For discriminator conditioning, we build on the projection method~\cite{miyato2018cgans} commonly used in modern GANs~\cite{brock2018large,karras2020training} defined as 
$f(x,y;\theta) = y^{T}V\phi(x;\theta_{\Phi}) + \psi(\phi(x;\theta_{\Phi}); \theta_{\Psi}),$ 
where $x$ is the network input, $y$ is the condition, $f(x,y;\theta)$ is the pre-activation discriminator output, $\theta = \{V, \theta_{\Phi}, \theta_{\Psi}\}$ are learned parameters such that $V$ is the embedding matrix of $y$, $\phi(x, \theta_{\Phi})$ is the network output given $x$ prior to conditioning, and $\psi(., \theta_{\Psi})$ is a scalar function of $\phi(x, \theta_{\Phi})$. However, this formulation extends only to either categorical or continuous attributes and does not apply to both types of conditioning, rendering it inadmissible to neuroimaging settings where we are simultaneously interested in attributes such as age (continuous) and disease (often categorical). Fortunately, under mild assumptions of conditional independence of the continuous and categorical attributes given the input, we find that similar analysis to~\cite{miyato2018cgans} factorizes cleanly into: $f(x,y;\theta) = y_{cat}^{T}V_{cat}\phi(x;\theta_{\Phi}) + y_{con}^{T}V_{con}\phi(x;\theta_{\Phi}) + \psi(\phi(x;\theta_{\Phi}); \theta_{\Psi})$,
where the $cat$ and $con$ subscripts indicate categorical and continuous attributes, respectively.
\begin{figure*}[t]
    \centering
    \includegraphics[width=\textwidth]{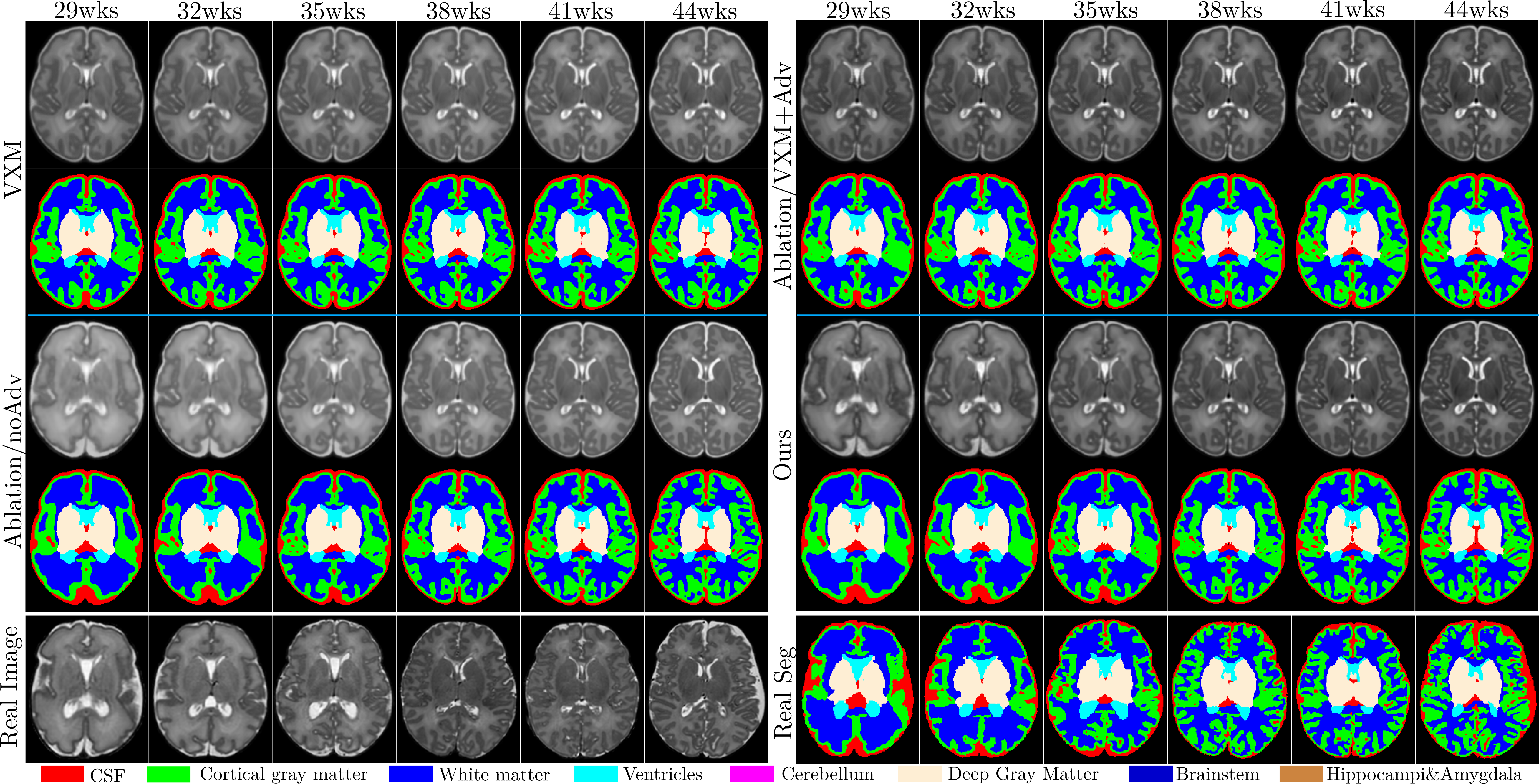}
    \caption{\textbf{Age-conditional dHCP templates} alongside template segmentations obtained by~\cite{makropoulos2014automatic}. Representative real images and segmentations are visualized in the bottom row.}
    \label{fig:dhcp-cond}
\end{figure*}
\begin{figure*}[t]
    \includegraphics[width=\textwidth]{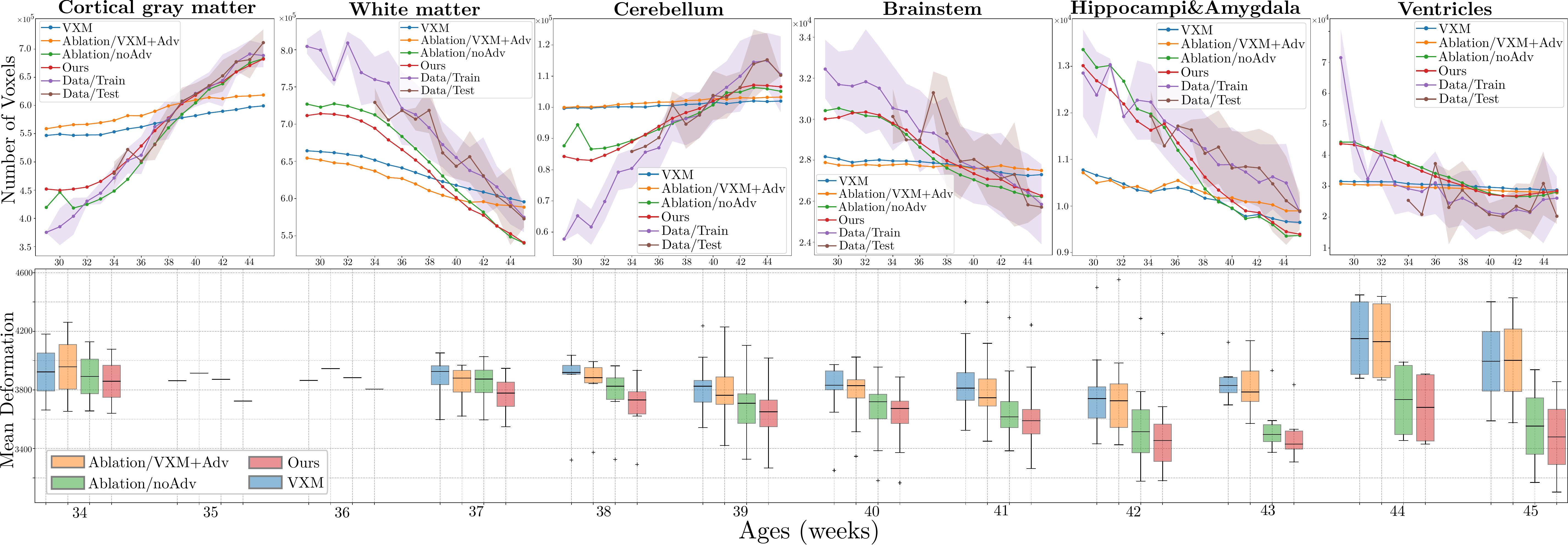}
    \caption{\textbf{Top:} volumetric trends of dHCP template segmentations for all methods overlaid upon the volumetric trends for the underlying train (purple) and test (brown) sets. \textbf{Bottom:} Mean deformation norms to held-out test data (lower is better) for all conditional methods.}
    \label{fig:dhcp-cond-quant}
\end{figure*}

\textbf{Loss Functions.} We define our objective function in the unconditional setting, with straightforward extensions to the conditional scenario.
The generator uses a three-part objective including an image matching term, penalties encouraging deformation smoothness and centrality, and an adversarial term encouraging realism in the moved images. For matching, we use a squared localized normalized cross-correlation (LNCC) objective, following standard medical image analysis rationale of requiring intensity standardization in local windows~\cite{AVANTS20112033}. 
For deformation regularization, we follow~\cite{dalca2019learning} and employ $Reg(\varphi) = \lambda_1 \| \bar{u} \|_{2}^2  + \lambda_2 \sum_{p \in \Omega} \| \nabla u(p) \|_{2}^2 + \lambda_3 \sum_{p \in \Omega} \| u(p) \|_{2}^2$ over voxels $p$,
where $u$ indicates spatial displacement such that $\varphi = Id + u$ and $\bar{u} = \frac{1}{n} \sum_{p \in \Omega} u(p)$ is a moving average of estimated spatial displacements. Briefly, the first term leads to small deformations across the entirety of the dataset, whereas the second and third encourage smooth and small individual deformations, respectively. The adversarial terms used to train generator and discriminator networks correspond to the least-squares GAN~\cite{Mao_2017_ICCV} objective, chosen for its relative stability. The overall generator loss can be summarized as $L = L_{LNCC} + \lambda_{reg}Reg(\varphi) + \lambda_{GAN}L_{GAN}$, where we use $\lambda_{GAN} = 0.1$ and $\lambda_{reg} = [\lambda_1, \lambda_2, \lambda_3] = [1, 1, 0.01]$ as in~\cite{dalca2019learning}. For the discriminator, we employ several forms of regularization, as detailed below.

\textbf{GAN Stabilization.} Generative adversarial training stabilizes and improves with lower image resolutions, higher batch sizes, bigger networks, and higher sample sizes~\cite{brock2018large}. However, the opposite arises in neuroimaging as images are larger volumes, GPU memory limits training configurations to low batch sizes and small networks, and sample sizes in medical imaging studies are often only a few hundred scans, and thus necessitate careful stabilization. We enforce a 1-Lipschitz constraint on both networks with spectral normalization~\cite{miyato2018spectral} on every layer, which has been shown to stabilize training and improve gradient feedback to the generator~\cite{Chu2020Smoothness}. We further use the $R_1$ gradient penalty~\cite{pmlr-v80-mescheder18a} on the discriminator which strongly stabilizes GAN training, defined as $R_{1} = \frac{\gamma}{2} \mathbb{E}_{x \sim P_{real}}[\|\nabla D(x)\|_{2}^{2}$ where $\gamma$ is the penalty weight, $P_{real}$ is the real distribution, and $D$ is the discriminator. As discriminator overfitting on limited data is a key cause of GAN instability, we further use differentiable augmentations~\cite{karras2020training,tran2020data,zhao2020differentiable,zhao2020image} on \textit{both} real and synthesized images when training the discriminator. We sample random translations for all datasets and further sample from the dihedral $D_4$ and (a subset of) $D_{4h}$ groups for 2D images and 3D volumes, respectively. Interestingly, brightness/contrast/saturation discriminator augmentations lead to training collapse for neuroimaging datasets, but were found to improve training on a 2D RGB face dataset.

\section{Experiments}
\subsection{Datasets}
\begin{figure*}[!t]
    \centering
    \includegraphics[width=\textwidth]{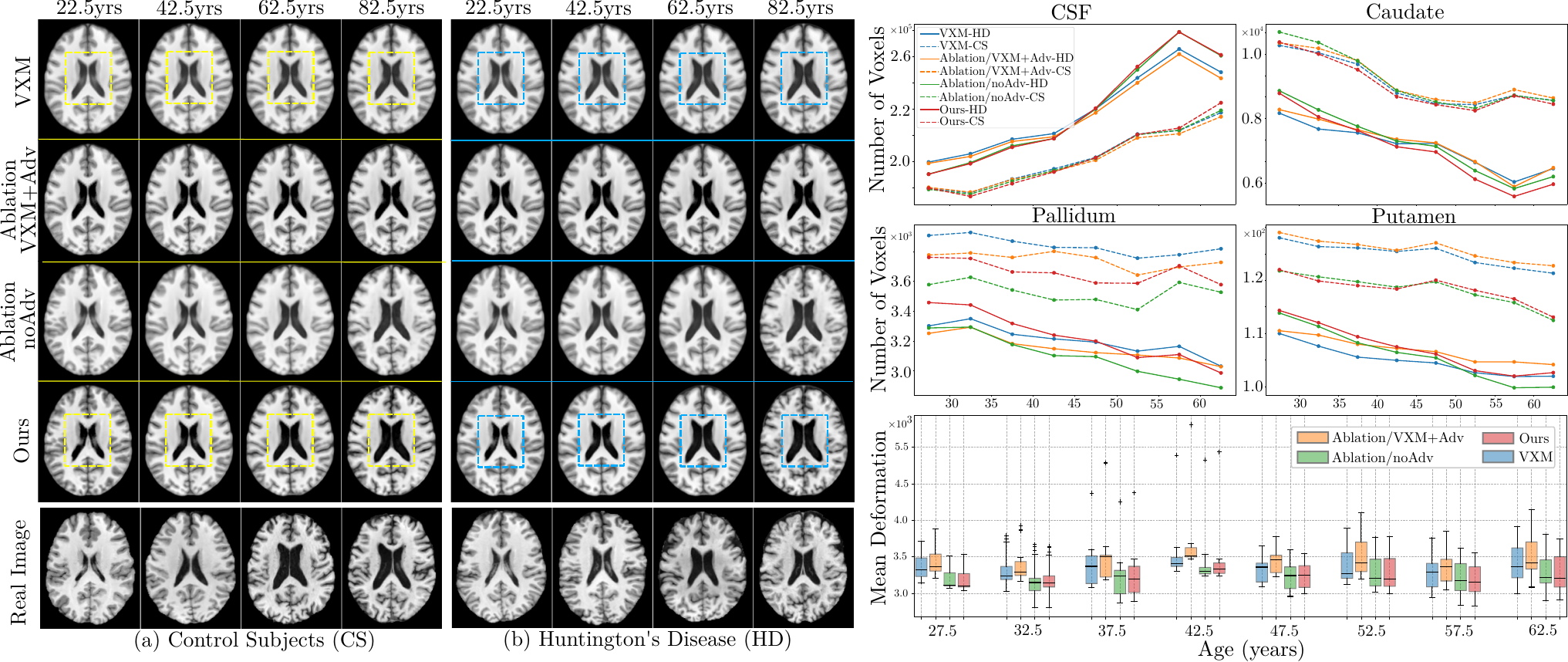}
    \caption{\textbf{Left:} Age and cohort-conditional templates for Predict-HD with representative real images visualized in the bottom row. \textbf{Top-right:} Inter-cohort volumetric trends from segmented templates for structures of interest for Huntington's disease. \textbf{Bottom-right:} Mean deformation norms to held-out test data (lower is better) for all conditional methods.}
    \label{fig:phd-cond}
\end{figure*}

\textbf{dHCP.} The developing human connectome project (dHCP) provides a dataset of newborns imaged near birth with gestational ages ranging from 29-45 weeks and thus displaying rapid week-to-week structural changes~\cite{hughes2017dedicated}. Spatiotemporal template estimation on dHCP is challenging as premature birth presents decreased cortical folding alongside increased incidences of hemorrhages, hyper-intensities, and lesions~\cite{molnar2013brain}. For age-conditioned template construction, we use all 558 T2w MR images from dHCP release 2 preprocessed and segmented by methodologies described in~\cite{makropoulos2018developing}. 
Images are affine-aligned to a constructed affine template and split at the subject-level (accounting for twins and repeat scans), resulting in 458, 15, and 85 scans for training, validation, and testing, respectively.

\textbf{Predict-HD.} We use a longitudinal multi-center and multi-scanner database of healthy controls and subjects with Huntington's disease (HD)~\cite{biglan2013refining,paulsen2008detection}. HD is a (typically) adult-onset progressive neurodegenerative disease impairing motor control and cognition~\cite{walker2007huntington} which substantially alters brain morphology. We build templates conditioned on age and the presence of the HD genetic mutation. We use 1117 T1w MR images from 388 individuals affine-aligned to MNI~\cite{fonov2011unbiased}. Image preprocessing is described in~\cite{paulsen2014prediction} and image segmentation was performed semi-automatically with labels corresponding to the Neuromorphometrics template~\cite{neuromorphometrics}. We use 897, 30, 190 images for training, validation, and testing, split at the subject level.

\textbf{FFHQ-Aging.} Face images have been used as experimental vehicles to analyze various qualitative aspects of template construction~\cite{avants2010optimal}. We use FFHQ-Aging~\cite{orel2020lifespan}, a database of $70,000$ real-world face images providing labels corresponding to (binned) age, gender, and the presence of glasses. FFHQ~\cite{karras2019style} captures significantly higher variation in terms of age, head pose, and accessories (e.g., hats and costumes) as compared to datasets such as CelebA~\cite{liu2018large} and is thus a significant challenge. We resize the training images to $128 \times 128$, and use age, gender, and the presence of glasses as input conditions. FFHQ-Aging is a challenging dataset, as topological changes (e.g., mouths open or closed) render diffeomorphisms to be a severely limited class of transformations for such images.

\begin{table*}[!t]
\caption{Quantitative evaluations on neuroimaging data for all methods including average Dice scores to test data, norms of accumulated moving average deformations over the course of training, entropy focus criteria (EFC), average Jacobian determinant to test data, and average deformation norms to test data. All deep network methods result in comparable Dice and Jacobian determinants, with our method (\texttt{Ours}) demonstrating improvements over baseline \texttt{VXM} in essential template qualities such as sharpness and deformation centrality.}
\centering
\begin{tabular}{@{}ccp{2.935cm}cccccc@{}}
\toprule
\multicolumn{2}{c}{\textbf{Setting}} & \textbf{Method} & \textbf{Avg Dice} ($\uparrow$) & \textbf{$\|$Mov. Def$\|$} ($\downarrow$) & \textbf{EFC} ($\downarrow$) & \textbf{Avg} $|J_{\varphi}|$ & \textbf{Avg $\|$Def$\|$} ($\downarrow$) \\ \midrule
\parbox[t]{2mm}{\multirow{5}{*}{\rotatebox[origin=c]{90}{Unconditional}}} & \parbox[t]{2mm}{\multirow{5}{*}{\rotatebox[origin=c]{90}{Predict-HD}}} & ANTs SyGN & $0.75 \pm 0.01$ & - & $0.882$ & $1.0000 \pm 0.0000$ & $4601 \pm 542$ \\
 & & VXM & $0.76 \pm 0.02$ & $238.83$ & $0.872$ & $0.9987 \pm 0.0016$ & $4029 \pm 296$ \\
 & & Ablation/VXM+Adv & $0.77 \pm 0.02$ & $209.42$ & $0.868$ & $0.9986 \pm 0.0014$ & $3919 \pm 289$ \\
 & & Ablation/noAdv & $0.77 \pm 0.02$ & $271.00$ & $0.864$ & $0.9979 \pm 0.0015$ & $4096 \pm 315$ \\
 & & Ours & $0.77 \pm 0.02$ & $224.68$ & $0.863$ & $0.9987 \pm 0.0013$ & $3872 \pm 291$ \\
\hline
\parbox[t]{2mm}{\multirow{5}{*}{\rotatebox[origin=c]{90}{Unconditional}}} & \parbox[t]{2mm}{\multirow{5}{*}{\rotatebox[origin=c]{90}{dHCP}}} & ANTs SyGN & $0.86 \pm 0.01$ & - & $0.872$ & $1.0000 \pm 0.0000$ & $2687 \pm 345$ \\
& & VXM & $0.88 \pm 0.01$ & $251.48$ & $0.894$ & $0.9992 \pm 0.0013$ & $3883 \pm 239$ \\
& & Ablation/VXM+Adv & $0.88 \pm 0.01$ & $261.36$ & $0.872$ & $0.9993 \pm 0.0014$ & $4029 \pm 234$ \\
& & Ablation/noAdv & $0.88 \pm 0.01$ & $285.55$ & $0.894$ & $0.9980 \pm 0.0013$ & $3987 \pm 257$ \\
& & Ours & $0.87 \pm 0.01$ & $250.27$ & $0.871$ & $0.9987 \pm 0.0012$ & $3815 \pm 232$ \\
\hline
\parbox[t]{2mm}{\multirow{4}{*}{\rotatebox[origin=c]{90}{Conditional}}} & \parbox[t]{2mm}{\multirow{4}{*}{\rotatebox[origin=c]{90}{Predict-HD}}} & VXM & $0.75 \pm 0.02$ & $221.68$ & $0.871 \pm 0.004$ & $0.9987 \pm 0.0013$ & $3867 \pm 252$ \\
& & Ablation/VXM+Adv & $0.75 \pm 0.02$ & $208.61$ & $0.863\pm0.004$ & $0.9987 \pm 0.0014$ & $3975 \pm 286$ \\
& & Ablation/noAdv & $0.75 \pm 0.02$ & $199.39$ & $0.874 \pm 0.005$ & $0.9987 \pm 0.0013$ & $3755 \pm 253$ \\
& & Ours & $0.75 \pm 0.02$ & $175.01$ & $0.863 \pm 0.007$ & $0.9991 \pm 0.0013$ & $3756 \pm 258$ \\
\hline
\parbox[t]{2mm}{\multirow{4}{*}{\rotatebox[origin=c]{90}{Conditional}}} & \parbox[t]{2mm}{\multirow{4}{*}{\rotatebox[origin=c]{90}{dHCP}}} & VXM & $0.87 \pm 0.01$ & $240.79$ & $0.901 \pm 0.006$ & $1.0001 \pm 0.0013$ & $3844 \pm 230$ \\
& & Ablation/VXM+Adv & $0.87 \pm 0.01$ & $215.90$ & $0.887 \pm 0.009$ & $0.9991 \pm 0.0012$ & $3823 \pm 239$ \\
& &  Ablation/noAdv & $0.86 \pm 0.03$ & $177.81$ & $0.903 \pm 0.007$ & $0.9990 \pm 0.0015$ & $3666 \pm 237$ \\
& & Ours & $0.87 \pm 0.01$ & $167.62$ & $0.885 \pm 0.009$ & $0.9994 \pm 0.0013$ & $3609 \pm 226$ \\ \bottomrule
\end{tabular}
\label{table:quant}
\end{table*}

\subsection{Baselines and Evaluation Strategies} \label{subsec:baselines_eval}
\textbf{Baselines \& Ablations.} We first compare with the widely-used unconditional template construction algorithm \verb|SyGN|~\cite{avants2010optimal} implemented in the \verb|ANTs| library~\cite{tustison2020antsx}. We then perform comparisons with a deep network for conditional and unconditional template estimation (\verb|VXM|~\cite{dalca2019learning}) trained under its original objective. To isolate our core differences from \texttt{VXM}, we use the same registration network for all settings. We use ablated variants to investigate whether adding a discriminator network to the original framework (\texttt{Ablation/VXM+Adv}) or whether training our architecture under only a regularized registration cost without a discriminator (\texttt{Ablation/noAdv}) yield similar improvements to our combined framework with both architectural changes and discriminator networks (\texttt{Ours}). As \texttt{Ablation/noAdv} is an ablation of \texttt{Ours}, it retains spectral normalization in the template generation branch, which may unnecessarily hamper its performance when trained without an adversarial cost. We do not compare with conventionally-optimized \textit{spatiotemporal} template construction methods as, to our knowledge, there are none that generically apply across diverse image sets (i.e., neonatal T2 MRI, adult T1 MRI, and RGB faces), account for arbitrary covariates, and typically require significant computational resources and domain knowledge.

\textbf{Evaluation.} Constructed templates are difficult to evaluate, as competing properties are often desired. For example, weak deformation regularization enables exact matching of templates to target images at the cost of generating anatomically-impossible deformation magnitudes and irregularities, whereas strong regularization provides smaller and more \textit{central} deformations but produces poor alignment~\cite{yeo2008sharpness}. We posit that preferable templates simultaneously present increased sharpness, accurate alignment, and small and smooth deformations to the target population.

We follow standard methods for quantifying template/MRI sharpness~\cite{esteban2017plos,joshi2004unbiased,liu2015low,rajashekar2020data} with the entropy focus criterion~\cite{efc}. To assess registration quality and centrality, we follow the evaluation protocols defined in~\cite{dalca2019learning} on held-out test data, including: (1) average Dice coefficients for segmentation labels deformed from the template to the target, corresponding to registration accuracy; (2) mean determinant of the Jacobian matrix $J_{\varphi}(p)$ of the deformation field $\varphi$ over voxels $p$ with $|J_{\varphi}(p)| \leq 0$ indicating local folding of the deformation field and $|J_{\varphi}(p)| \sim 1$ corresponding to smooth local deformation; (3) average deformation norms to the target images, with lower values indicating improved template centrality given equivalent registration accuracy. Finally, we estimate the norm of the moving average of deformations accumulated over training iterations, with lower values corresponding to increased centrality.

The considered datasets present significant gaps between training and test set age distributions and hence quantitative registration evaluations are performed on a subset of the overall age range. For Dice evaluation, Predict-HD template segmentation followed~\cite{dalca2019learning} and dHCP templates were segmented with~\cite{makropoulos2014automatic}. See the appendices for further details.

\begin{figure*}[!ht]
    \centering
    \includegraphics[width=\textwidth]{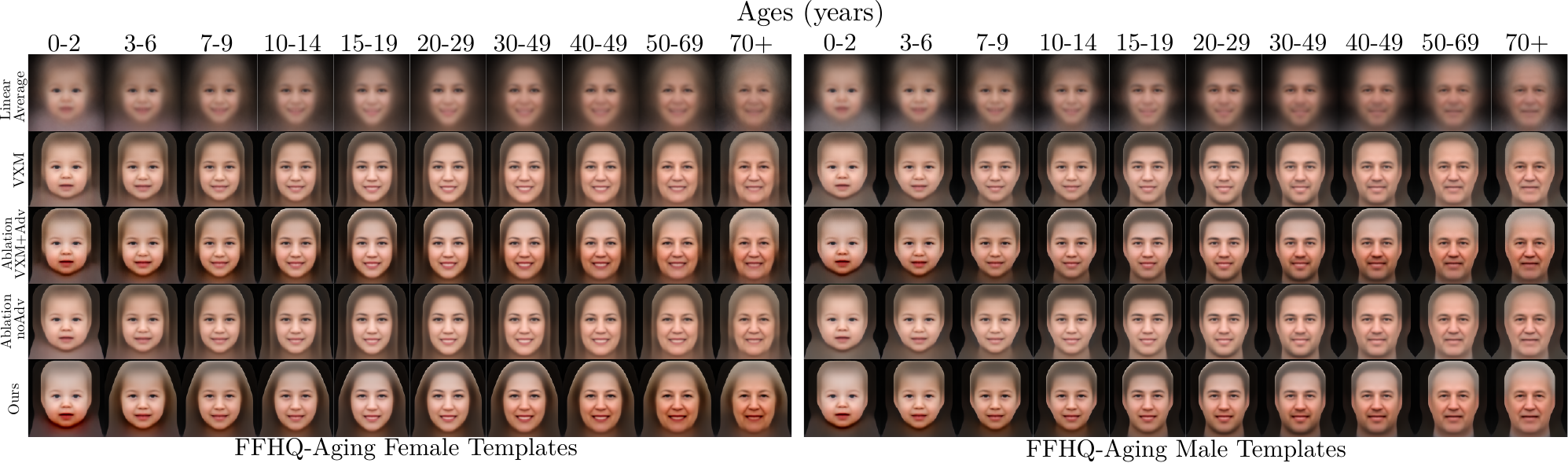}
    \caption{\textbf{Age and cohort-conditional FFHQ templates} showing qualitatively improved perceptual fidelity with our framework.}
    \label{fig:ffhq-cond}
\end{figure*}

\subsection{Implementation Details}
We use a batch size of 1 for 3D neuroimages and a batch size of 32 for 2D planar images. As all datasets considered have highly imbalanced age distributions, we sample minority time-points at higher frequencies when training conditionally. We do not make dataset-specific hyperparameter choices beyond $R_1$ regularization for GAN stability, where we use $\gamma=5\times10^{-4}, 10^{-3}$, and $5\times10^{-3}$ for Predict-HD, dHCP, and FFHQ-Aging, respectively, corresponding to their respective stabilization requirements. For fair comparison, the hyperparameters and architectures pertaining specifically to registration were matched to the suggested settings for \texttt{VXM} for all deep networks. All architectures and remaining design choices are detailed in the appendices.

\subsection{Results and Analysis}

\textbf{Figure \ref{fig:uncond-atlas}} shows unconditional templates for dHCP and Predict-HD, highlighting that the adversarial approach yields more anatomically-accurate templates. For example, in the sagittal view  of the Predict-HD templates (bottom row), we observe anatomical structures more clearly within the red insets when using adversarial regularizers, whereas prior methods are unable to do so. 
\verb|SyGN| is unable to resolve structures which display high-frequency deformations due to its reliance on averaging aligned intensities and shapes. 

For unconditional templates, we observe subtle differences between the proposed method and its reduction \texttt{Ablation/VXM+Adv}, with either approach having distinct advantages depending on their intended use. \texttt{Ablation/VXM+Adv} trains faster as it does not use 3D convolutions in its template generation branch. Conversely, \texttt{Ours} removes subtle checkerboard patterns generated by previous methods. Interestingly, unconditional \texttt{Ablation/noAdv} produces stronger moving average deformation magnitudes than other deep network approaches (Table \ref{table:quant}), suggesting that the adversarial objective is necessary for this setting to obtain optimal results. However, subsequent analysis of \textit{conditional} template estimation reveals significant differences between these approaches. 

\textbf{Figure \ref{fig:dhcp-cond}} provides sample age-conditional templates alongside anatomical segmentations for dHCP. Methods that use input-concatenation architectures for template generation (\texttt{VXM} and \texttt{Ablation/VXM+Adv}) underfit the rapid week-to-week developmental changes in this dataset, whereas conditionally-modulated architectures (\texttt{Ablation/noAdv} and \texttt{Ours}) can generate templates which closely follow the underlying trends in the data.
\textbf{Figure \ref{fig:dhcp-cond-quant}} further shows that the conditionally-modulated models better represent volumetric changes in the held-out test set (top row) while showing improved centrality (bottom row). In comparison to \texttt{Ablation/noAdv}, the complete framework (\texttt{Ours}) better fits spatiotemporal image contrast, is sharper, and shows mildly improved centrality (\textbf{Table} \ref{table:quant}). We underscore that the training and test image volumes correspond to the \textit{affine pre-aligned} data and thus reflect relative volumes in the affine template space.

We observe similar trends in \textbf{Figure \ref{fig:phd-cond}} (left) for Predict-HD. While age and cohort-conditional templates generated by \texttt{VXM} and \texttt{Ablation/VXM+Adv} do show subtle geometric variation, stronger spatiotemporal changes and dataset-similarity are observed with our complete framework (compare changes in ventricles within the dashed boxes). Analogous improvements in test-set deformation magnitude for Predict-HD are shown in the boxplots (bottom-right). Volume trends of segmented templates for regions pertinent to Huntington's disease are shown in the top-right and follow expected trends such as larger ventricular volumes and smaller basal ganglia volumes in the group with the Huntington's mutation as compared to the controls. 

\textbf{Figure \ref{fig:ffhq-cond}} illustrates qualitative FFHQ-Aging templates. Row 1 shows the conditional linear averages of the training set. As above, methods trained without adversarial losses (rows 2 and 4) correctly learn spatiotemporal changes while methods trained with them (rows 3 and 5) present stronger appearance variation and yield improved template perception, e.g. by removing border artefacts in the male cohort. \texttt{Ours} further removes border artefacts from the female cohort and increases shape variation and perceptual fidelity (bottom-left row). Further analysis of FFHQ-Aging template construction is presented in the appendices.

\textbf{Table \ref{table:quant}} summarizes quantitative results. All methods achieve comparable Dice coefficients and produce smooth deformations ($|J_{\varphi}| \sim 1$). However, the proposed techniques generally yield improved sharpness (lower entropy focus criteria) and improved centrality (lower deformation norms) while showing equivalent registration performance, indicating that the constructed templates are more barycentric representations of the data. We stress that Dice coefficients between unconditional and conditional settings cannot be directly compared as the template segmentations were obtained via different approaches. Finally, EFC interpretation requires care. While the numerical differences are subtle, EFC range is more restricted. For example, Gaussian filtering ($\sigma = 1$) of the Predict-HD \textit{unconditional} template from our method only increases EFC from $0.86$ to $0.88$, indicating that smaller changes are meaningful. Conditional EFC indicate that using a discriminator significantly improves sharpness across scan age ($p < 10^{-5}$ between \verb|VXM| and \verb|Ours| for both datasets), with trends best interpreted via the temporal EFC plots given in the appendices.

\section{Conclusions}
Explicit methods for the construction of conditional deformable templates using stochastic gradient descent and deep networks are powerful tools for the efficient and flexible modeling of image populations with arbitrary covariates. In this work, we take a generative synthesis approach towards explicit template estimation by constructing a framework which enables training on datasets challenging for generative adversarial networks. The resulting templates are sharp and easy to delineate for domain-experts, are more representative of the underlying demographics, and closely follow typical development in both neonatal MRI of developing pre-term neonates and adult MRI sampled across typical lifespans with and without neurodegeneration. Finally, while this work is motivated from the perspective of neuroimaging, it applies to generic imaging modalities. 

\section*{Acknowledgements}

This work was supported by NIH grants NIBIB R01EB021391, 1R01HD088125-01A1, 1R01DA038215-01A1, U01 NS082086, 1R01AG064027, and NS050568. Part of the HPC resources used for this research were provided by NSF grant MRI-1229185. Original and processed Predict-HD image data and segmentations were provided by the University of Iowa as part of a collaborative NIH grant U01 NS082086. dHCP results were obtained using data made available from the Developing Human Connectome Project funded by the European Research Council under the European Union’s Seventh Framework Programme (FP/2007-2013) / ERC Grant Agreement no. [319456]”.

{\small
\bibliographystyle{ieee_fullname}
\bibliography{references}
}

\onecolumn
\begin{appendices}

\section{Supplementary Results}
\textbf{Supplementary animations} illustrating several conditional spatiotemporal experiments are available at \url{https://www.neeldey.com/deformable-templates}.

\begin{figure}[!h]
\centering
\includegraphics[width=\textwidth]{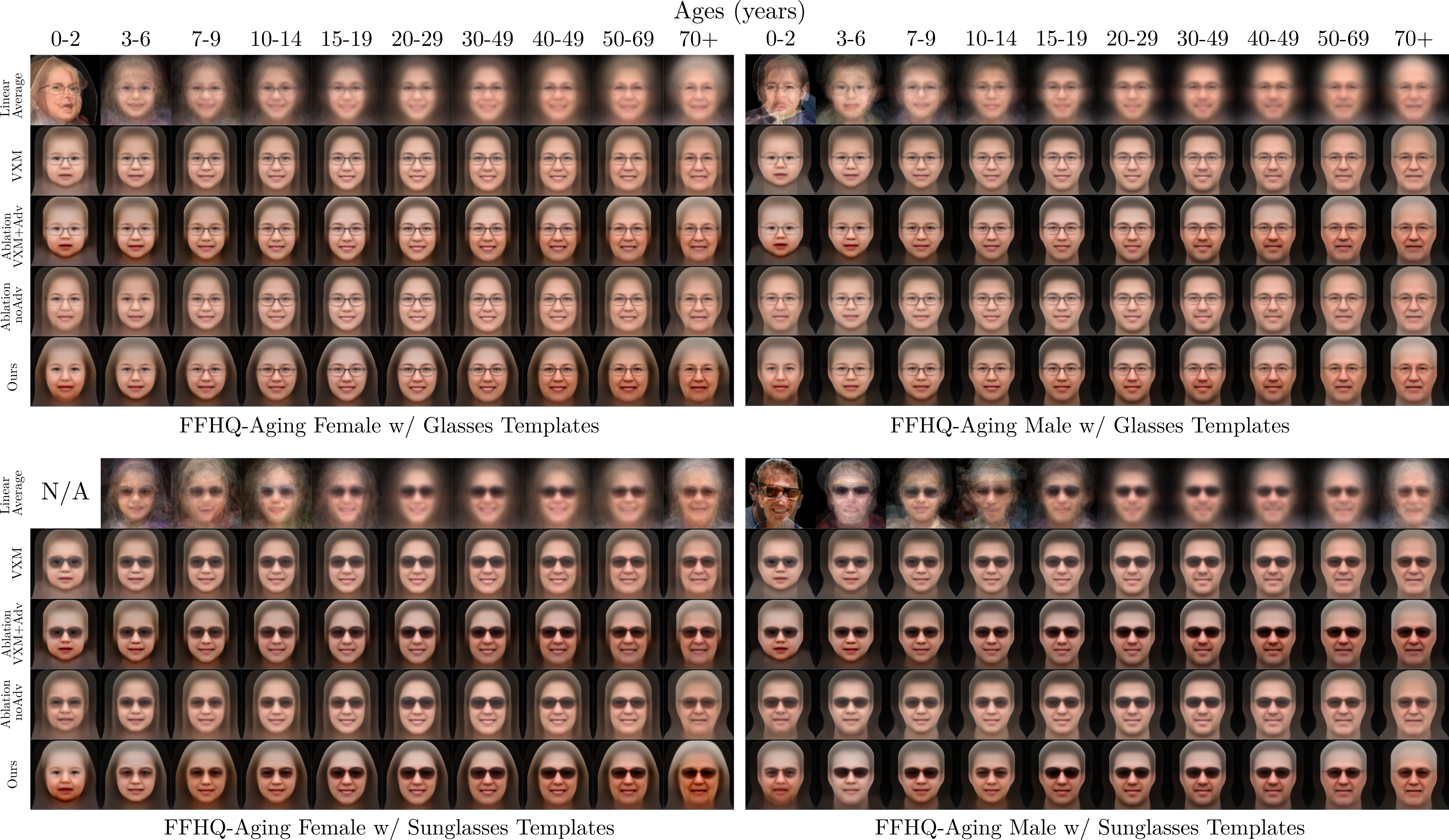}
\caption{FFHQ-Aging age and cohort conditional templates with normal glasses (top) and sunglasses (bottom). For ages 7 and older, all methods produce plausible conditional templates, with \texttt{Ours} removing border effects and increasing shape and appearance variability. Significant label noise and highly limited sample sizes are apparent for ages 0-2 within the ``glasses" label and for ages 0-6 within the ``sunglasses" labels. For example, only two images exist within the training set for the male/with sunglasses/0-2 years old FFHQ-aging class with both images displaying \textit{adults} with sunglasses and not infants (as can be seen from the corresponding linear average). As a result, methods using FiLM~\cite{perez2018film} (\texttt{Ablation/noAdv} and \texttt{Ours}) produce more adult-like templates in those age ranges. We speculate the results come from the increased data fitting capacity of FiLM. Interestingly, methods which do not use FiLM (\texttt{Ablation/VXM+Adv} and \texttt{VXM}) produce more plausible age-conditioned templates when all of the data for a category is mislabeled. This phenomenon arising from significant label noise requires future investigation. }
\label{fig:ffhq-glasses}
\end{figure}

\begin{figure}[!hptb]
\centering
\includegraphics[width=0.95\textwidth]{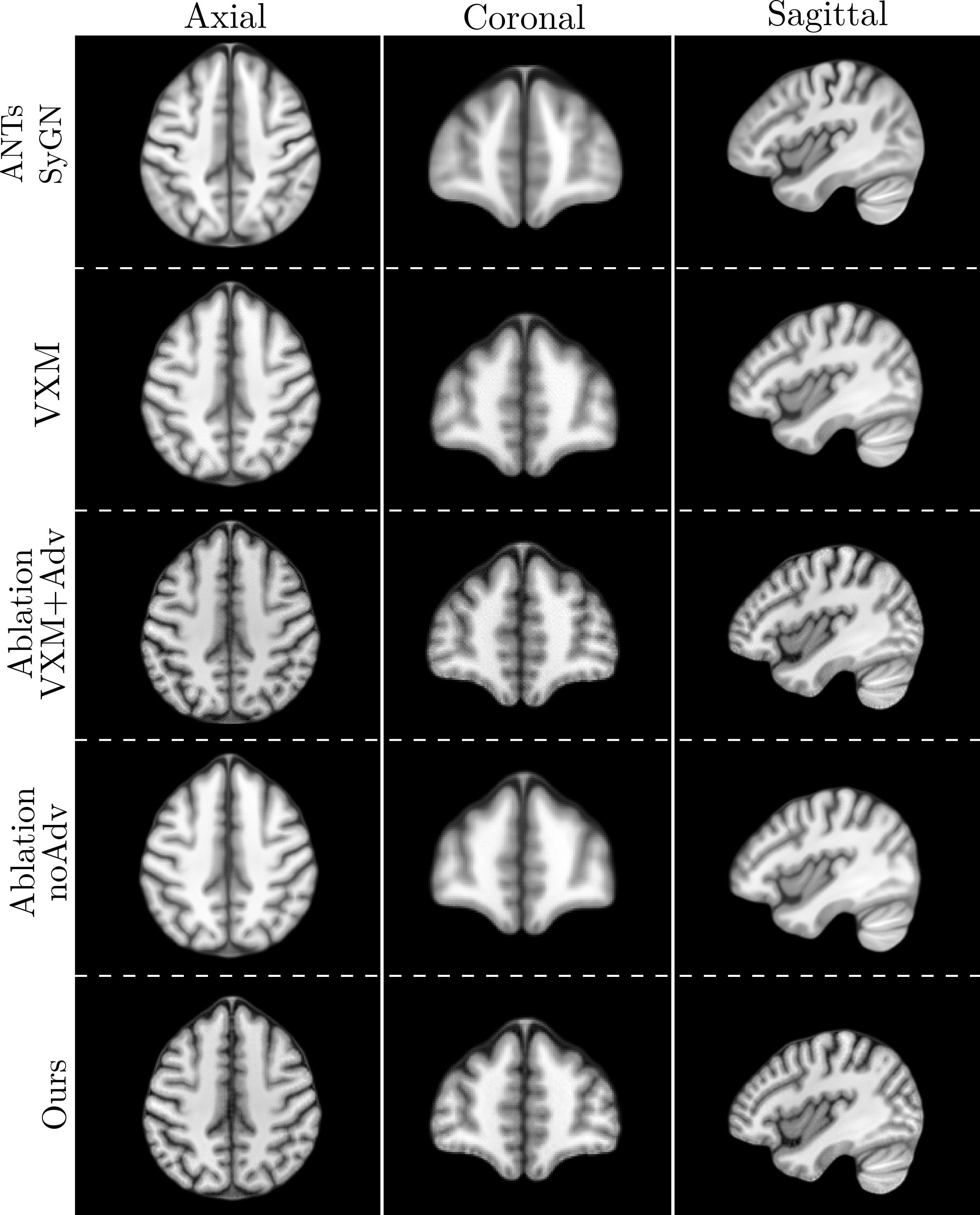}
\label{fig:uncond}
\caption{Additional 2D views of unconditional 3D template construction on Predict-HD from all four methods. Methods using a discriminator (\texttt{Ablation/VXM+Adv} and \texttt{Ours}) exhibit increased sharpness, cortical folding detail, and tissue contrast. \texttt{Ours} improves on \texttt{Ablation/VXM+Adv} by removing subtle checkerboard artefacts, particularly visible in the coronal view.}
\end{figure}

\begin{figure}
    \centering
    \includegraphics[width=\textwidth]{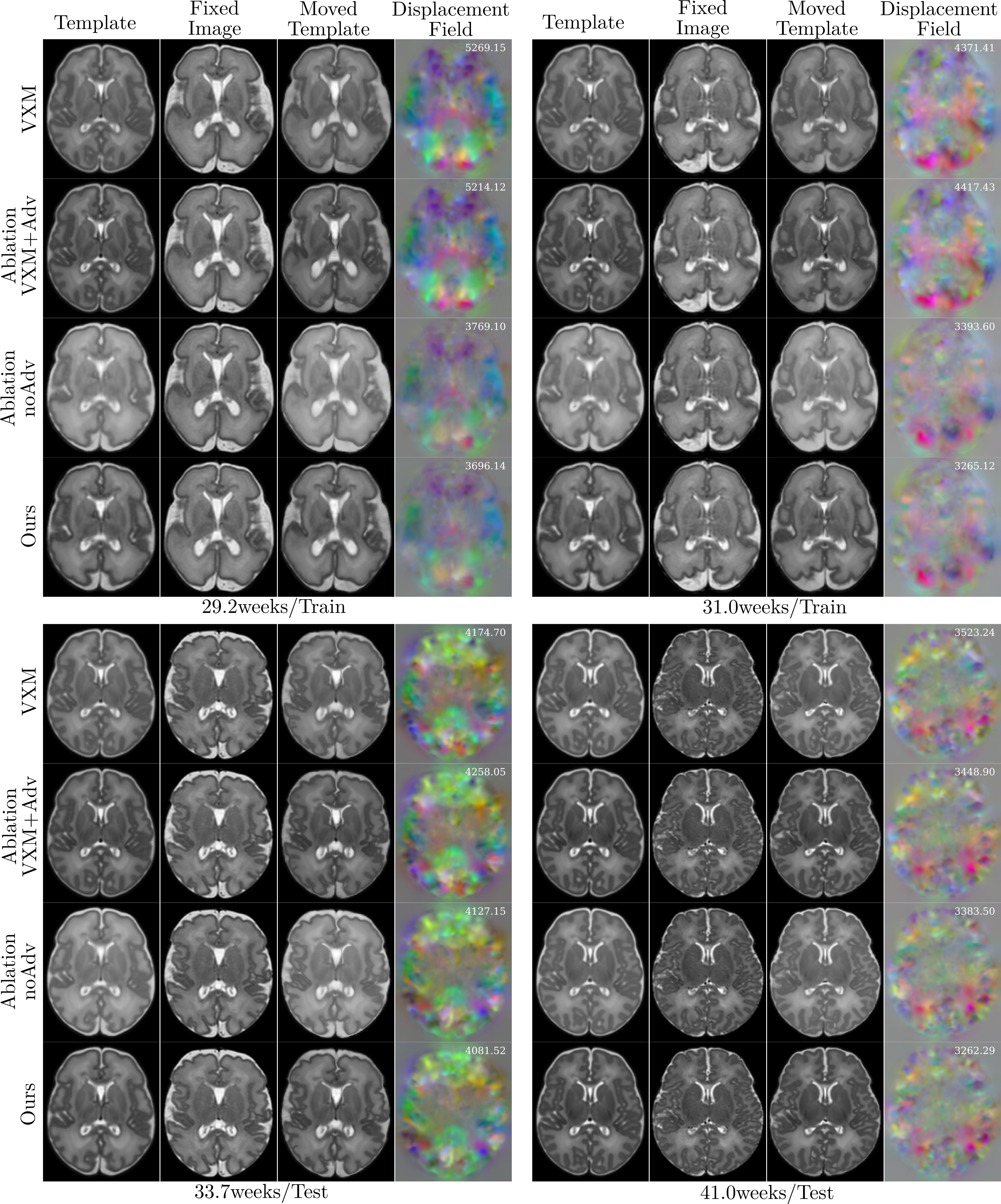}
    \caption{Example dHCP template-to-image registration results for all methods on training data (top subfigures) and held-out test data (bottom subfigures), with varying gestational ages. Deformation norms for the 3D displacement fields are annotated on the top-right. We visualize training set results in addition to test data as a large age range (29-31weeks) of interest is not present in the test set (See Figure~\ref{fig:data-distrib}). As our templates show higher condition (age) specificity, the deformations are smaller and more anatomically plausible as compared to baselines and ablations.}
    \label{fig:dhcp-reg}
\end{figure}

\begin{figure}
    \centering
    \includegraphics[width=\textwidth]{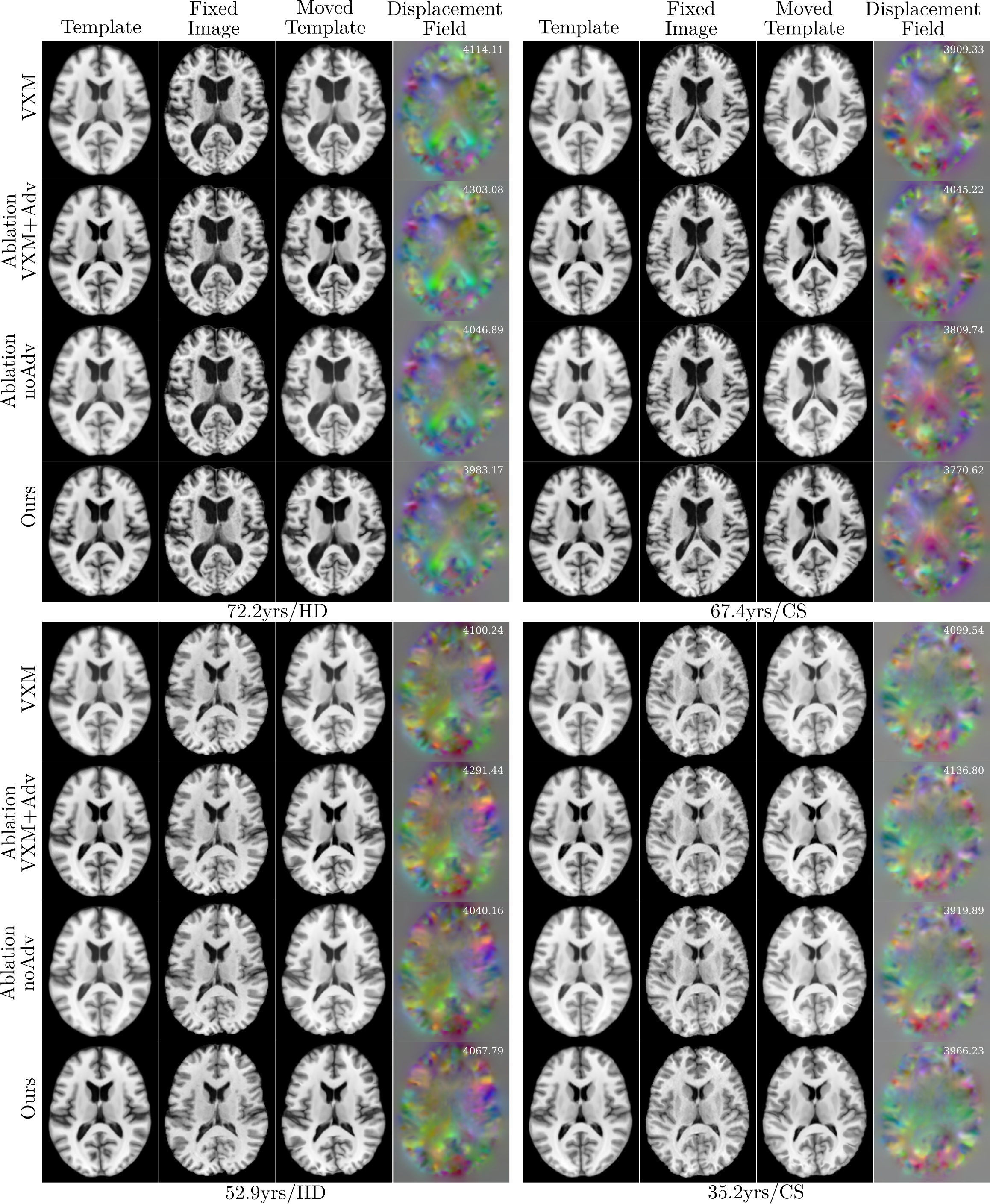}
    \caption{Example Predict-HD template-to-image registration results for all methods on held-out test data, with varying ages and cohorts. Deformation norms for the 3D displacement fields are annotated on the top-right. All methods produce comparable moved templates. However, ours yields smaller deformations as seen from the displacement fields (especially visible in 72.2yrs/HD and 67.4yrs/CS).}
    \label{fig:phd-reg}
\end{figure}

\begin{figure}[!htbp]
\centering
\includegraphics[width=0.78\textwidth]{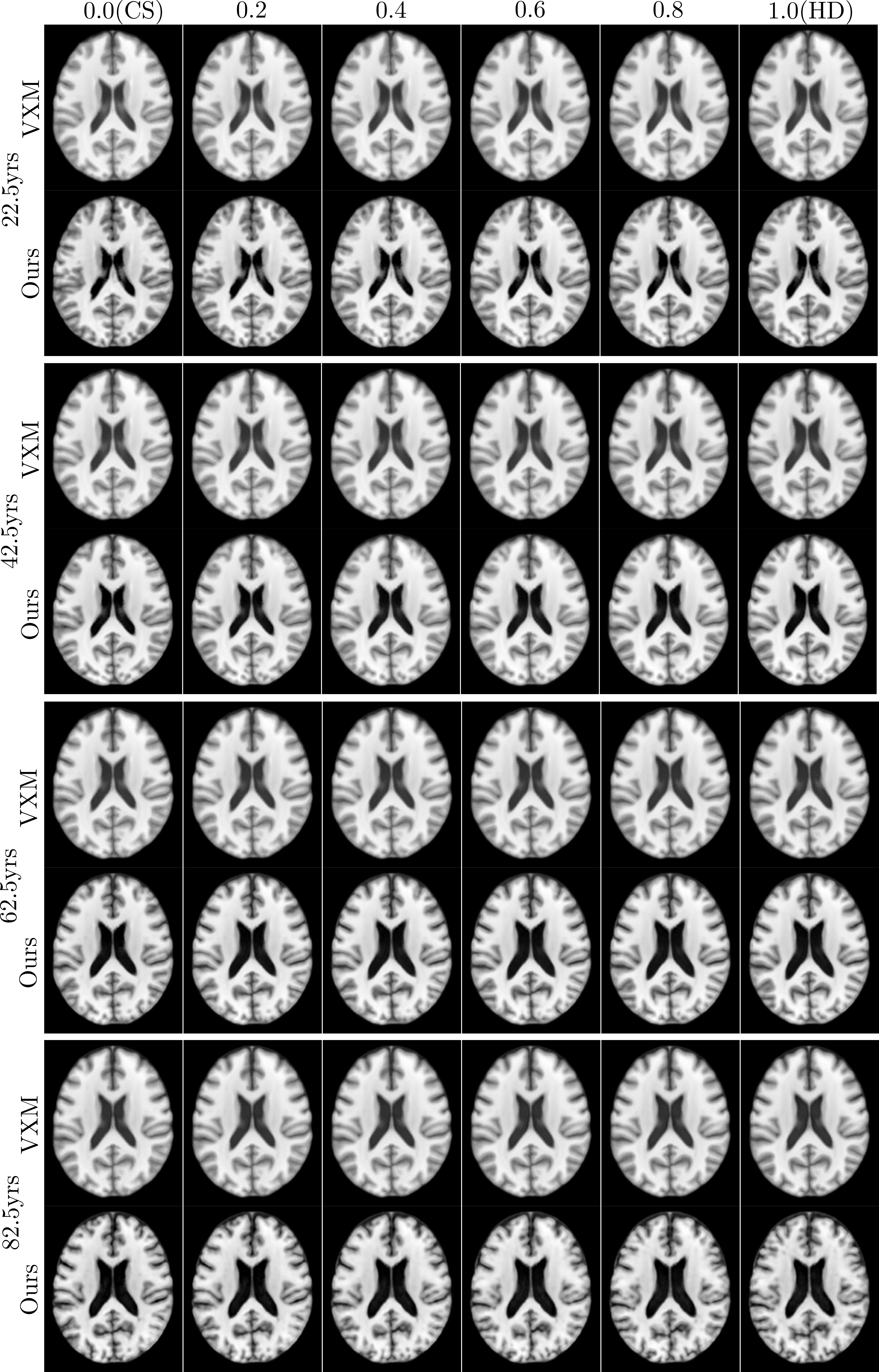}
\label{fig:CStoHDinterps}
\caption{Interpolations between control subjects (CS) and subjects with the Huntington's disease (HD) mutation ($[0,0.2,0.4,0.6,0.8,1]$), for fixed ages, obtained by linearly interpolating between one-hot attribute vectors. Both methods (\texttt{VXM} and \texttt{Ours}) achieve interpolations which match clinical expectations, e.g., with ventricles growing larger as the HD weight increases. \texttt{Ours} displays larger differences between CS and HD with correspondingly larger changes visible in the interpolations, as can be seen from the last row of the figure.}
\end{figure}

\begin{figure}[!htpb]
\centering
\includegraphics[width=0.9\textwidth]{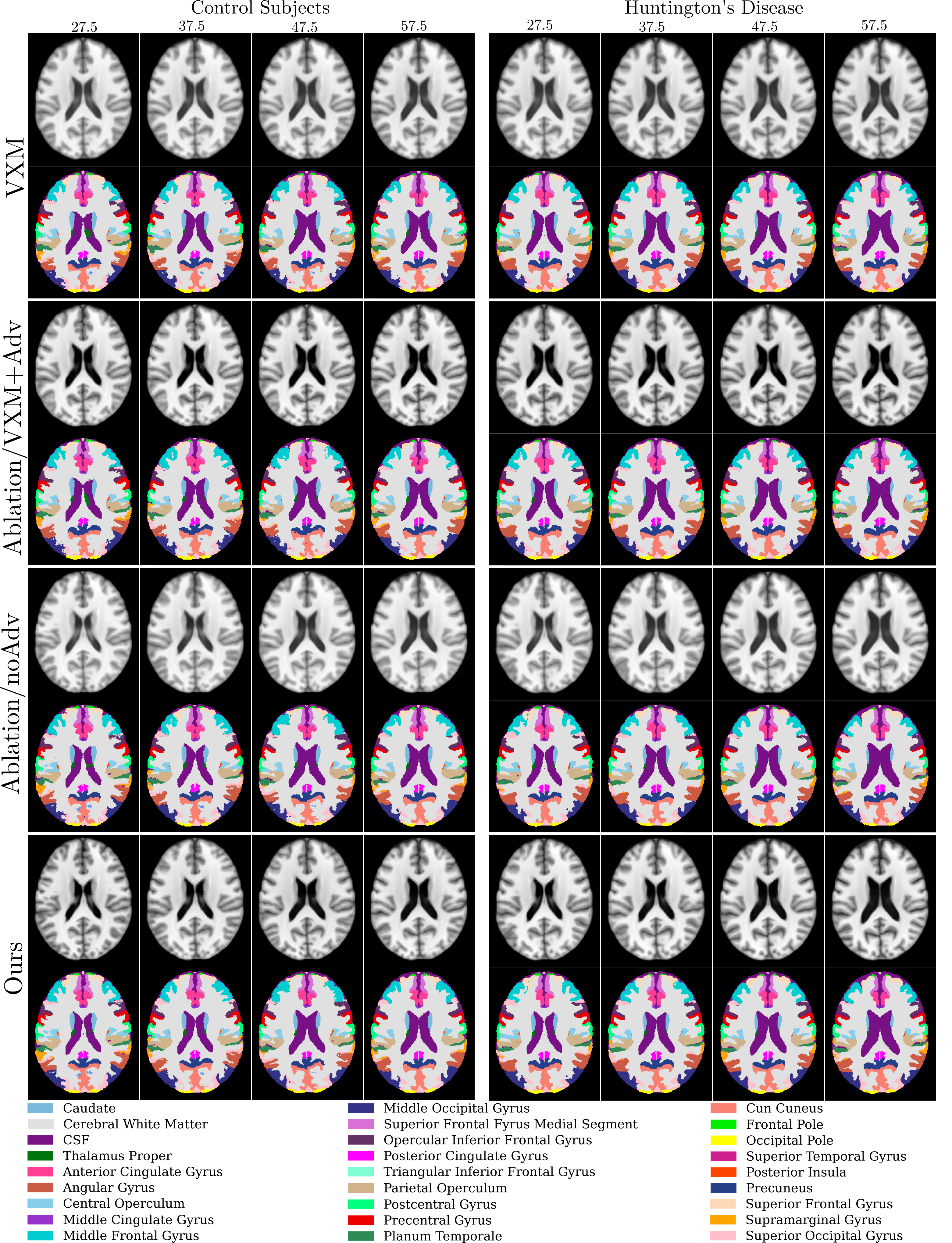}
\label{fig:phd-w-seg}
\caption{Example template segmentations for all methods generated by majority voting on inverse warped labels of training images. We emphasize that no segmentation labels are used in template construction or registration and that these segmentations are used only for Dice coefficient evaluation and temporal volume trends. 
}
\end{figure}

\begin{figure}[!ht]
\centering
\includegraphics[width=\textwidth]{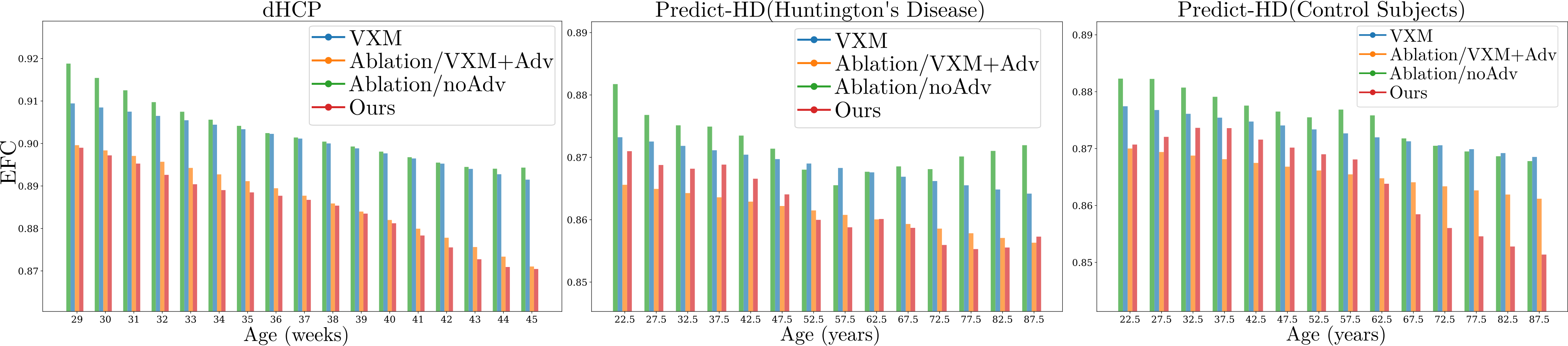}
\label{fig:efc}
\caption{Temporal Entropy Focus Criteria (EFC, lower is better) for conditional templates on the dHCP (left), Predict-HD/Huntington's Disease (center), and Predict-HD/Control Subjects (right). In all cases, methods using a discriminator (\texttt{Ablation/VXM+Adv} and \texttt{Ours}) achiever lower EFC over non-generative adversarial methods. These results should be interpreted in context as:
\newline (1) While \texttt{Ablation/VXM+Adv} and \texttt{Ours} achieve equivalent EFC/sharpness, \texttt{Ours} displays increased condition-specificity and centrality as shown in the experiments in the primary text.
\newline (2) Although commonly used to evaluate unconditional template sharpness, EFC is a heuristic surrogate for image sharpness and can fluctuate with varying structure. As \texttt{Ablation/noAdv} and \texttt{Ours} show strong structural changes temporally, their temporal trends show higher variability as compared to techniques which present smaller structural changes (\texttt{Ablation/VXM+Adv} and \texttt{VXM}). As a result, EFC should be compared across methods at individual timepoints.}
\end{figure}

\begin{figure}
    \centering
    \includegraphics[width=\textwidth]{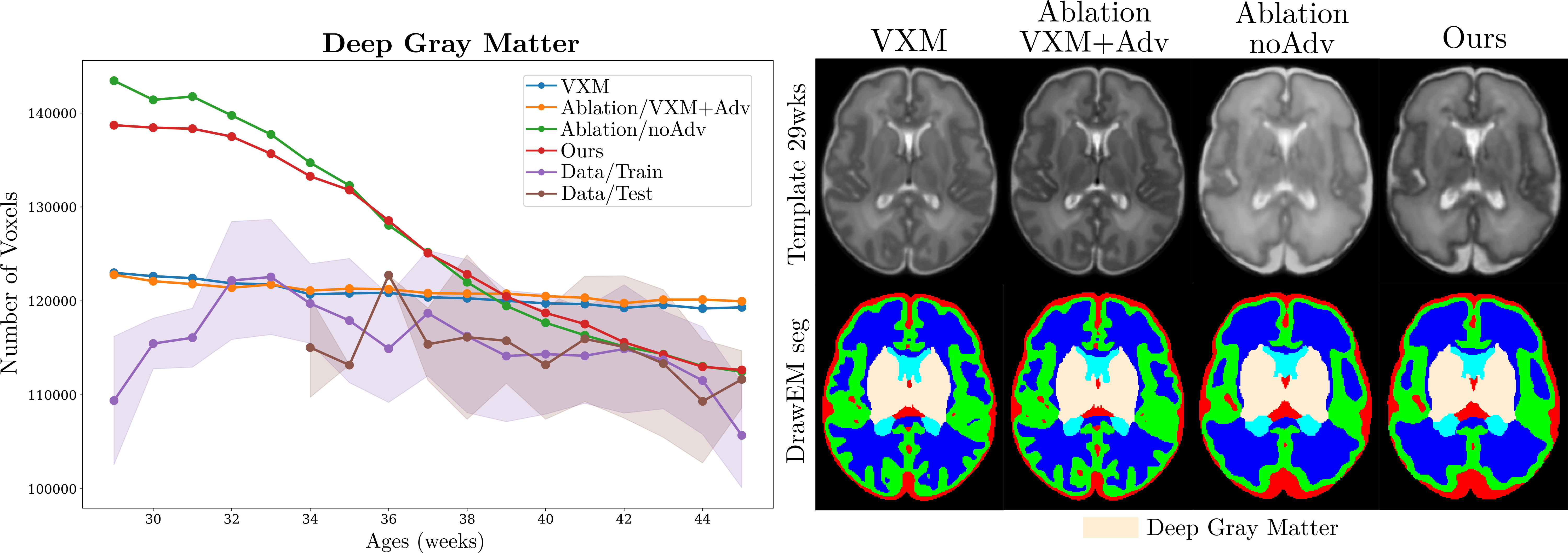}
    \caption{Negative results for dHCP conditional template segmentation for the Deep Gray Matter (dGM) label. \texttt{DrawEM}~\cite{makropoulos2014automatic} (the tool used for dHCP template segmentation) with its default parameters overestimates dGM volume on the templates sampled at younger timepoints by \texttt{Ablation/noAdv} and \texttt{Ours}. For example, on the right, we show the generated templates from all methods at 29 weeks gestational age, with their \texttt{DrawEM} segmentation results below. While \texttt{Ablation/noAdv} and \texttt{Ours} produce more anatomically plausible templates compared to  \texttt{VXM} and \texttt{Ablation/VXM+Adv}, the segmentation algorithm overestimates dGM volume. All other labels better match the underlying volume trends on the real data as shown in Figure 4 of the main text. In future work, careful tuning of DrawEM parameters on validation data may resolve this dGM overestimation.}
    \label{fig:dhcp-deepgm}
\end{figure}

\newpage 
\section{Experimental Details}
\subsection{Data Preparation}
\begin{figure}[!ht]
\centering
\includegraphics[width=\textwidth]{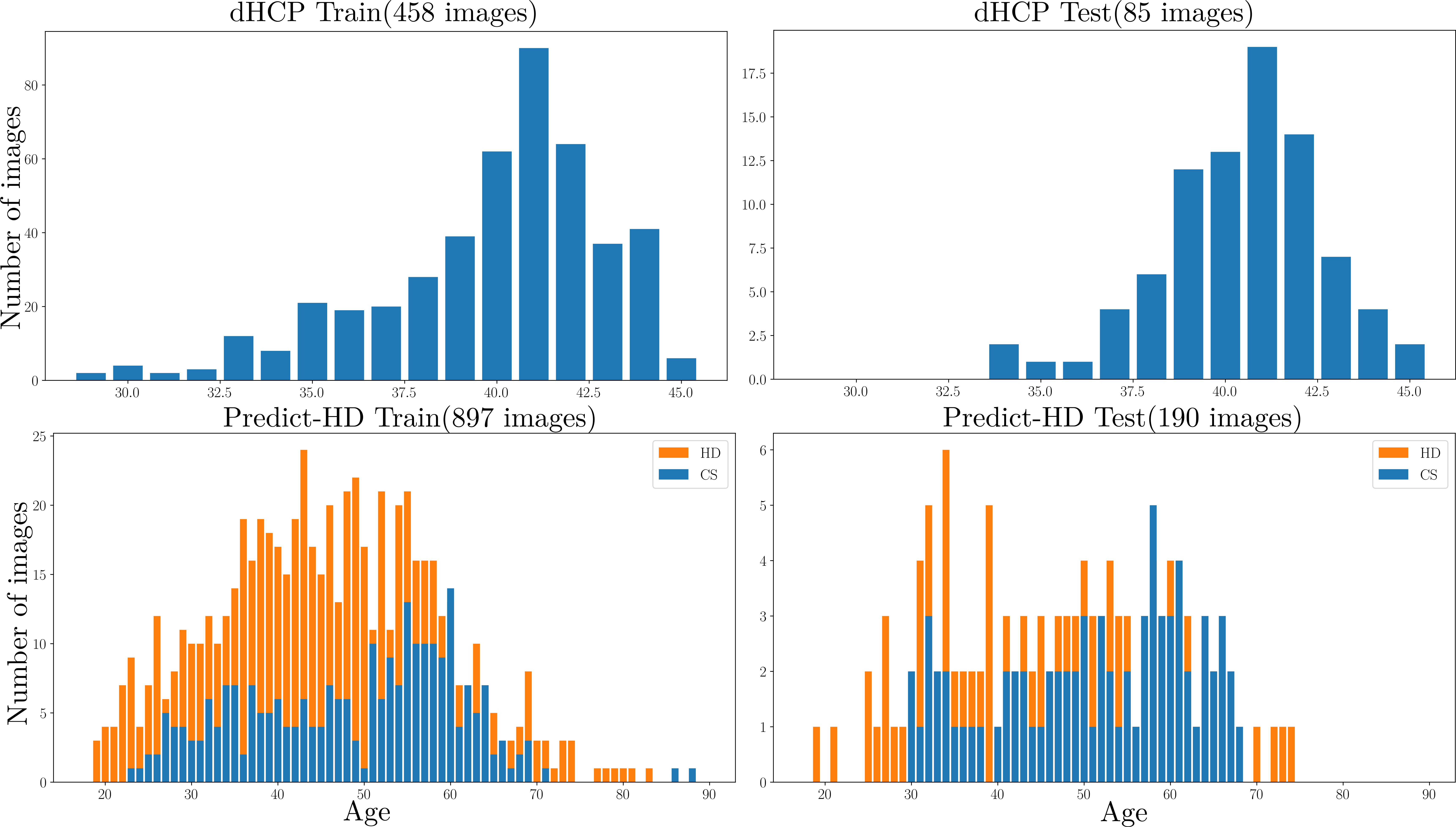}
\label{fig:data-distrib}
\caption{Histograms of sample size vs. scan age for both dHCP (top row) and Predict-HD (bottom row), for the constructed training sets (left column) and test sets (right column). Both datasets present both a significant gap between training and test sets in terms of scan age sampling.}
\end{figure}

All foreground/brain extraction is performed by thresholding provided segmentation labels. All neuroimages are cropped to a central field-of-view of resolution $160 \times 192 \times 160$. We obtain all linear averages required for the neuroimaging experiments using voxel-wise averages of a 100 randomly chosen training scans. All 60,000 training images are used to compute the linear average for FFHQ (while we visualize group-wise $L_2$ barycenters in row 1 of Figure \ref{fig:ffhq-glasses} for comparison, our framework uses the overall $L_2$ barycenter for training).

\subsubsection{Predict-HD}
Predict-HD provides longitudinal scans from 388 individuals with and without the Huntington's Disease (HD) genetic condition. As imaging was performed across several distinct scanning sites, the images present highly heterogeneous appearance. All T1 images were bias corrected and segmented using procedures described in~\cite{paulsen2014prediction}. Prior to learning nonlinear deformable registration, we affinely register all T1 images to MNI~\cite{fonov2011unbiased}, thus resampling them to $1\times1\times1 mm^{3}$. Out of 1121 images, 4 either failed affine alignment or had missing covariates and were discarded. We use 897, 30, and 190 images for training, validation, and testing, respectively, split at the subject level. In the context of this study, we do not currently consider longitudinal subject-specific effects in our conditional template estimation. 

To compute template-to-image registration accuracy via Dice coefficient evaluation, we follow the template segmentation protocols outlined in~\cite{dalca2019learning}. Briefly, we select training scans within the ages of 25 and 65 years old wherein we have sufficient sample sizes for both cohorts and for both train and test sets. Accordingly, our Dice coefficient evaluation is only conducted on held-out test subjects between the ages of 25 and 65 (176 out of 190). The images are split into 5-year-wide age bins with a single template sampled at the center of the bin (i.e., a 52.5 year old HD template for HD subjects with ages between 50 and 55). For each cohort, all training segmentations within a bin are inverse warped to the bin-specific template, followed by majority voting on the labels to obtain the template segmentation for that age-bin and group. Unconditional template segmentations are performed via the same procedure, without the need for binning time points. In the future, other label fusion methods accounting for local intensities can be incorporated~\cite{wang2012multi}.

\subsubsection{dHCP}
Release 2 of the developing human connectome project (dHCP) was pre-processed by a specialized pipeline for neonatal image analysis~\cite{makropoulos2018developing} including steps such as motion-correction, super-resolution (from $0.8\times0.8\times1.6 mm^{3}$ to $0.5\times0.5\times0.5 mm^{3}$), bias correction, brain extraction, and segmentation~\cite{makropoulos2014automatic}. For GPU memory, we crop to a central field-of-view and minimally resize images from $0.5 \times 0.5 \times 0.5 mm^3$ voxel resolution to $0.6132 \times 0.6257 \times 0.6572 mm^3$ for a final image size of $160 \times 192 \times 160$. For training, validation, and testing, we first assign all twins and repeat scans to the training set to prevent test set leakage and randomly hold-out a 100 scans from the remainder (15 for validation, 85 for testing), resulting in 458 training images. We construct an affine template for the training set with \verb|ANTs| to which every scan is affinely aligned. 

To generate segmentations for conditional templates generated by all methods for use in computing registration accuracy via Dice coefficients and analyzing volumetric trends of anatomical regions-of-interest, we use \verb|DrawEM| ~\cite{makropoulos2014automatic}. Briefly, \verb|DrawEM| is a multi-atlas EM-segmentation pipeline based on the neonatal ALBERT templates~\cite{gousias2012magnetic} using normalized mutual information based image registration. This is in contrast to the majority voting template segmentation procedure for Predict-HD above and ~\cite{dalca2019learning}. \verb|DrawEM| segmentation was performed instead of majority voting as several time points have very limited sample sizes not suitable for majority voting when using regularized registration, leading to qualitatively inaccurate template segmentations. We find that \verb|DrawEM| produces sufficiently accurate segmentations on templates produced by all methods (as shown in Figure 3 of the main text, see Figure~\ref{fig:dhcp-deepgm} in the supplementary material for a counter-example). Unconditional template segmentation was performed following~\cite{dalca2019learning}.

\subsubsection{FFHQ-Aging}
FFHQ-Aging~\cite{orel2020lifespan} annotates images in the FFHQ~\cite{karras2019style} human face dataset. These annotations include genders, ages (in 10 age bins), head pose (pitch, roll, and yaw), type of glasses (no glasses, normal glasses, sunglasses), eye occlusion scores, and segmentation labels (obtained by a DeepLabV3~\cite{chen2017rethinking} model pre-trained on CelebAMask-HQ). For simplicity, we only use the categorical attributes, leaving head pose and eye-occlusion conditioning to future work. We train all models on the FFHQ training set of 60,000 images (out of 70,000). As is common~\cite{avants2010optimal}, we restrict ourselves to qualitative evaluations for face templates. Importantly, we note that categorical template conditions for human faces are quite coarse as attributes such as gender are not purely categorical. Further, we note that the data set is skewed towards lighter skin tones (as evidenced by the linear averages of training images visualized in Supplementary Figure \ref{fig:ffhq-glasses}), which is consequently reflected in the synthesized templates from all methods. In future work, more careful modeling and diverse data collection protocols may work towards ameliorating these issues.

\subsection{Additional Implementation Details}
\paragraph{Design choices and hyperparameters.} Architectures are given in Table \ref{arch1}. All estimated templates for neuroimaging experiments are masked by a foreground mask during training for all methods to suppress commonly occurring background artefacts. The foreground mask was obtained by thresholding a linear average of training images. Reflection padding was used instead of zero-padding for all methods as it led to slightly fewer checkerboard artefacts. LeakyReLU slopes were set to 0.2. A window of 100 updates is used for the moving average deformation penalty $\bar{u}$ for all datasets. For condition vectors $z$, we encode age as a continuous attribute (for the neuroimaging where we have access to continuous age values) divided by the maximum age in the dataset, and categorical attributes as one-hot vectors. We find that not rescaling continuous attributes in $z$ can lead to discriminator instability. Weight decay was applied on the linear projections from the FiLM embedding to the individual layers with weight $10^{-5}$ for neuroimages and $10^{-6}$ for FFHQ-Aging.

We choose the stationary velocity field (SVF~\cite{arsigny2006logeuc,ASHBURNER200795}) framework primarily for its speed and ease of implementation and note that other frameworks such as LDDMM~\cite{beg2005lddmm} can also be used. The integration over time $t \in \left[ 0, 1 \right]$ is in practice implemented for all methods with five \textit{scaling and squaring} layers which have been shown to produce smooth diffeomorphic displacement fields~\cite{arsigny2006logeuc,dalca2019diffeo}. While all training is performed on full resolution 3D volumes, velocity and displacement fields are estimated at half-resolution and then linearly scaled up during training as in~\cite{dalca2019learning}. This resizing has an implicit smoothing effect. Implementations of spatial transformers and scaling and squaring layers are taken from the \verb|voxelmorph| library at \url{voxelmorph.mit.edu}.

For FFHQ-Aging, we make a few changes from the neuroimaging datasets. FFHQ-Aging provides ages in categorical form and are thus treated as one-hot representations. Linear averages for FFHQ-Aging were computed across the entire training dataset due to the high number of classes. As the dataset has a left-right head pose asymmetry (particularly pronounced in subclasses with few samples), we use horizontal reflection augmentation for all methods when training the template generation and registration sub-networks. We further use a penalty $\|I - I_{LR}\|_2^{2}$ with unit weight (where $I_{LR}$ indicates a left-right reflection of $I$) for all methods to encourage symmetric face templates.

\paragraph{Optimization Details.} As GAN training involves the joint optimization of two networks, the optimization parameters used in either network impacts training stability. The Adam~\cite{kingma2014adam} optimizer is used in all networks. For conditional dHCP and Predict-HD training, we adopt a two-time-scale-update-rule (TTUR~\cite{heusel2018gans}), with step size $\eta_G = 10^{-4}$, $\eta_D = 3 \times 10^{-4}$, using $\beta_1 = 0.0$, $\beta_2 = 0.9$ in both networks as is common in recent GAN works~\cite{brock2018large,Park_2019_CVPR}. For conditional FFHQ-Aging, we reduce $\eta_D$ to $2 \times 10^{-4}$ as additional stability was needed for highly challenging face registration. Unconditional template optimization was found to be amenable to mild amounts of momentum and was performed with step-size $\eta = 10^{-4}$, $\beta_1 = 0.5$, and $\beta_2 = 0.999$ used in both generator and discriminator for faster convergence. We note that momentum is theoretically contraindicated for $R_1$ gradient penalty on the discriminator but we did not find this to be an issue in practice. The non-GAN baselines (\verb|VXM| and \verb|Ablation/noAdv|) were trained with the same strategies to enable valid comparisons.

\paragraph{ANTs SyGN parameters.} We use the \verb|antsMultivariateTemplateConstruction2.sh| script provided by the ANTsX ecosystem~\cite{tustison2020antsx} which implements the SyGN algorithm from~\cite{avants2010optimal}. We use the default construction parameters, including the squared local normalized cross-correlation objective, four template updates, using a four-level registration pyramid with at $6\times, 4\times, 2\times$ downsampling for the first three resolutions, and $100 \times 100 \times 70 \times 20$ iterations per resolution. We turn off the default bias field correction and linear registration steps as these are performed during data pre-processing. Registrations between the estimated template and held-out test images were performed with the same registration parameters. We leave the default Laplacian sharpening on for all comparisons.

\paragraph{Miscellaneous Experimental Details.} All networks are implemented in \verb|TensorFlow 2.2| and trained on a single NVIDIA V100 GPU. As the GAN frameworks (\verb|Ours| and \verb|Ablation/VXM+Adv|) require concurrent optimization of two 3D networks, we found 16 GB vRAM neccessary for training. All entropy focus criteria are calculated within a common brain mask for the dataset.

\section{Projection Discriminator}
We use the inner product-based framework presented in \cite{miyato2018cgans} who observe that the optimum for the standard adversarial loss can be written as (equation 2 of \cite{miyato2018cgans}):
\begin{equation*}
    f^{\ast}(x, y) = log\Big(\frac{q(y|x)q(x)}{p(y|x)p(x)}\Big) = log\Big(\frac{q(y|x)}{p(y|x)}\Big) + log\Big(\frac{q(x)}{p(x)}\Big) := r(y|x) + r(x)
\end{equation*}
where $x$ represents unconditional input, $y$ represents conditional information, and $q$ and $p$ are the real and synthesized data distributions, respectively. When we have conditioning $y = [y_{cat}, y_{con}]$ such that $y_{cat}$ is categorical and $y_{con}$ is continuous, assuming that they are conditionally independent given $x$, we obtain through simple modification:
\begin{align*}
    f^{\ast}(x, y) &= log\Big(\frac{q(y_{cat}, y_{con}|x)q(x)}{p(y_{cat}, y_{con}|x)p(x)}\Big) \\
    &= log\Big(\frac{q(y_{cat}, y_{con}|x)}{p(y_{cat}, y_{con}|x)}\Big) + log\Big(\frac{q(x)}{p(x)}\Big) \\
    &= log\Big(\frac{q(y_{cat}|x)q(y_{con}|x)}{p(y_{cat}|x) p(y_{con}|x)}\Big) + log\Big(\frac{q(x)}{p(x)}\Big) \\
    &= log\Big(\frac{q(y_{cat}|x)}{p(y_{cat}|x)}\Big) +
    log\Big(\frac{q(y_{con}|x)}{p(y_{con}|x)}\Big) + log\Big(\frac{q(x)}{p(x)}\Big) := r_{cat}(y|x) + r_{con}(y|x) + r(x),
\end{align*}
with the remaining analysis following \cite{miyato2018cgans} leading to the projection-discriminator expression given in the main text.

\begin{table}[t]
\parbox{.45\linewidth}{
\begin{center}
\begin{tabular}{c}
\toprule
\textbf{Template Generator} ($T$) \\
\toprule
Inputs: conditions $z \in \mathbb{R}^c$ \\
\hline
Embed $z \in \mathbb{R}^c$ into $\hat z \in \mathbb{R}^{64}$ using $C$\\
\hline
Learn Parameters $h \in \mathbb{R}^{80 \times 96 \times 80 \times 8}$ \\
FiLM($\hat{z}$) \\
\hline
ConvSN, 8 $\rightarrow$ 32 \\
\hline
$5 \times$ ResBlockSN, 32 $\rightarrow$ 32 \\
\hline
Upsample $2 \times$ trilinearly \\
\hline
ConvSN, $32 \rightarrow 8$, FiLM($\hat{z}$), LeakyReLU \\
\hline
ConvSN, $8 \rightarrow 8$, FiLM($\hat{z}$), LeakyReLU \\
\hline
ConvSN, $8 \rightarrow 8$, FiLM($\hat{z}$) \\
\hline
ConvSN, $8 \rightarrow 1$, FiLM($\hat{z}$), tanh \\
\hline
Add to average of training images for $T(\hat{z})$ \\
\bottomrule
\end{tabular}
\end{center}
}
\hfill
\parbox{.45\linewidth}{
\begin{center}
\begin{tabular}{c}
\toprule
\textbf{Registration Network} ($R$) \\
\toprule
Inputs: template $T(\hat{z})$; target $F$ \\
\hline
$h_0:$ Concatenate($T(\hat{z})$, $F$) \\
\hline
$h_1:$ Conv, Stride 2, 2 $\rightarrow$ 32, LeakyReLU \\
\hline
$h_2:$ Conv, Stride 2, 32 $\rightarrow$ 32, LeakyReLU \\
\hline
$h_3:$ Conv, Stride 2, 32 $\rightarrow$ 32, LeakyReLU \\
\hline
$h_4:$ Conv, Stride 2, 32 $\rightarrow$ 32, LeakyReLU \\
\hline
$h_5:$ Conv, 32 $\rightarrow$ 32, LeakyReLU \\
\hline
$h_6:$ Conv, 32 $\rightarrow$ 32, LeakyReLU, Up $2\times$, concat $h_3$ \\
\hline
$h_7:$ Conv, 32 $\rightarrow$ 32, LeakyReLU, Up $2\times$, concat $h_2$ \\
\hline
$h_8:$ Conv, 32 $\rightarrow$ 32, LeakyReLU, Up $2\times$, concat $h_1$ \\
\hline
$h_9:$ Conv, 32 $\rightarrow$ 32, LeakyReLU \\
\hline
$h_{10}:$ Conv, 32 $\rightarrow$ 32, LeakyReLU \\
\hline
$h_{11}:$ Conv, 32 $\rightarrow$ 16 \\
\hline
$v:$ ConvBlock, 16 $\rightarrow$ 3 \\
\hline
$\varphi:$ 5 $\times$ Scale and Square($v$) \\
\hline
$M(T(\hat{z})):$ STN($T(\hat{z}), \varphi$) \\
\bottomrule
\end{tabular}
\end{center}
}
\bigskip
\parbox{.45\linewidth}{
\begin{center}
\begin{tabular}{c}
\toprule
\textbf{Discriminator} ($D$) \\
\toprule
Inputs: image $x \in \mathbb{R}^{160 \times 192 \times 160}$; attributes $z \in \mathbb{R}^{c}$ \\
\hline
ConvSN, stride 2, $1 \rightarrow 64$, Leaky ReLU \\
\hline
ConvSN, stride 2, $64 \rightarrow 128$, Leaky ReLU \\
\hline
ConvSN, stride 2, $128 \rightarrow 256$, Leaky ReLU \\
\hline
ConvSN, stride 2, $256 \rightarrow 512$, Leaky ReLU \\
\hline
Conv, stride 1, $512 \rightarrow 64$ to $D'(x)$ \\
\hline
Projection($D'(x)$, $z$) \\
\bottomrule
\end{tabular}
\end{center}
}
\hfill
\parbox{.45\linewidth}{
\begin{center}
\begin{tabular}{c}
\toprule
\textbf{Embedding/FiLM Generator} ($C$) \\
\toprule
Inputs: attributes $z \in \mathbb{R}^{c}$ \\
\hline
Dense(64), LeakyReLU \\
\hline
Dense(64), LeakyReLU \\
\hline
Dense(64), LeakyReLU \\
\hline
Dense(64), LeakyReLU for $C(z)$ \\
\bottomrule
\end{tabular}
\end{center}
}
\caption{Architectures for Conditional Predict-HD and dHCP consisting of a template generator (\textbf{top left}), a registration network (\textbf{top right}), a discriminator network (\textbf{bottom left}) and a FiLM embedding generator (\textbf{bottom right}). \texttt{Conv} represents a $3\times3\times3$ convolutional layer (\texttt{ConvSN} indicates use of spectral normalization). A \texttt{ResBlockSN} consists of two blocks of sequential \texttt{ConvSN} and \texttt{LeakyReLU} layers with an additive skip connection. For unconditional template estimation, we do not use any FiLM layers. For FFHQ-Aging, we use the same architectures, only replacing the 32 per-layer filters with 64 in the template generator (due to the higher number of classes), using \texttt{ConvSN} instead of \texttt{Conv} in the generator and the penultimate layer of the discriminator, and reducing the channel multiplier in the discriminator from 64 to 32.}
\label{arch1}
\end{table}
\end{appendices}

\end{document}